\documentclass[10pt,twocolumn,letterpaper]{article}

\usepackage{iccv}
\usepackage{times}
\usepackage{epsfig}
\usepackage{graphicx}
\usepackage{algpseudocode}
\usepackage{algorithmicx,algorithm}
\usepackage{alltt}
\usepackage[dvipsnames]{xcolor,colortbl}
\usepackage{float}
\usepackage{subcaption}
\usepackage{amsmath}
\usepackage{amssymb}
\usepackage{booktabs}
\usepackage{color}
\usepackage[dvipsnames]{xcolor,colortbl}
\usepackage{svg}
\usepackage{multicol,multirow}
\usepackage{makecell}
\usepackage[font=small,labelfont=bf,tableposition=top]{caption}
\usepackage[pagebackref=true,breaklinks=true,letterpaper=true,colorlinks,bookmarks=false]{hyperref}

\iccvfinalcopy 

\def\iccvPaperID{1211} 

\ificcvfinal\fi

\begin{document}

\title{DOT: A Distillation-Oriented Trainer}

\author{Borui Zhao$^{1}$ \quad Quan Cui$^{2}$ \quad Renjie Song$^{1}$ \quad  Jiajun Liang$^{1}$ \\
\vspace{-10pt}\\
$^{1}$MEGVII Technology \quad $^{2}$Waseda University\\
{\tt\small zhaoborui.gm@gmail.com,  cui-quan@toki.waseda.jp,} \\ 
{\tt\small song.renjie@foxmail.com, liangjiajun@megvii.com} 
}

\ificcvfinal\thispagestyle{empty}\fi

\makeatletter
\let\@oldmaketitle\@maketitle
\renewcommand{\@maketitle}{\@oldmaketitle
  \centering
  \includegraphics[width=0.9\textwidth]{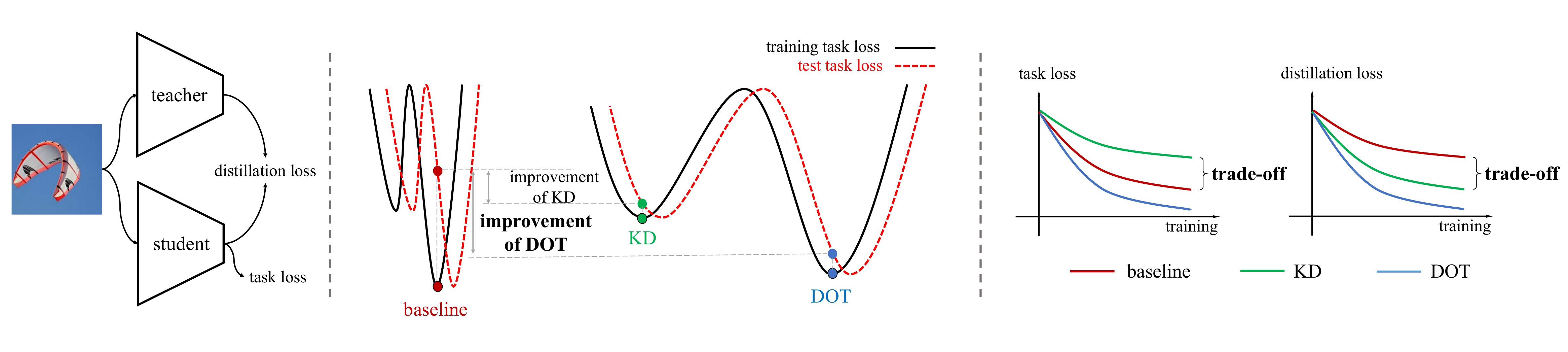}
    \captionof{figure}{\small{\textbf{\textit{Left}}: the framework of knowledge distillation~(KD). KD introduces an extra distillation loss, transferring knowledge from the teacher model. \textbf{\textit{Middle}}: a conceptual sketch of flat and sharp local minima~\cite{minima1,asymvalley}. The Y-axis denotes the loss value, and the X-axis the network parameters. The considerable sensitivity of the training loss at sharp minima damages the generalization on test data. In this paper, we discover that knowledge distillation~(KD) benefits the student baseline~(CE) \textit{with flatter minima} but unexpectedly \textit{limits the convergence}. See Figure~\ref{fig:fig2} and \ref{fig:fig2c}. \textbf{\textit{Right}}: the task and distillation loss dynamics. It suggests that introducing KD brings about a trade-off between the task and distillation losses. See Figure~\ref{fig:fig2c}. To address this \textit{trade-off} issue and achieve better performance, we propose  Distillation-Oriented Trainer~(DOT). Our DOT breaks the trade-off and leads the student to ideal minima of \textbf{both great flatness and convergence}.}}
    \label{fig:fig1} \bigskip} 
\makeatother

\maketitle

\begin{abstract}

\vspace{-10pt}

Knowledge distillation transfers knowledge from a large model to a small one via task and distillation losses. 
In this paper, we observe a \textbf{trade-off} between task and distillation losses, \ie, introducing distillation loss limits the convergence of task loss. 
We believe that the trade-off results from the \textit{insufficient} optimization of distillation loss. The reason is: The teacher has a lower task loss than the student, and a lower distillation loss drives the student more similar to the teacher, then a better-converged task loss could be obtained. 
To break the trade-off, we propose the Distillation-Oriented Trainer~(DOT). 
DOT separately considers gradients of task and distillation losses, then applies a larger momentum to distillation loss to accelerate its optimization. We empirically prove that DOT breaks the trade-off, \ie, both losses are sufficiently optimized.
Extensive experiments validate the superiority of DOT. Notably, DOT achieves a \textbf{+2.59\%} accuracy improvement on ImageNet-1k for the ResNet50-MobileNetV1 pair. Conclusively, DOT greatly benefits the student's optimization properties in terms of loss convergence and model generalization. Code will be made publicly available.
\end{abstract}


\section{Introduction}
Knowledge distillation~\cite{kd,survey,survey2,NIPS2017_e1e32e23,NEURIPS2021_892c91e0,NEURIPS2021_29c0c0ee,NEURIPS2021_018b59ce} has been proved to be an effective manner to transfer knowledge from a heavy~(teacher) model to a light~(student) one in a wide range of deep learning tasks~\cite{vgg,resnet,faster_rcnn,devlin2018bert,collobert2008unified}.
Novel learning algorithms have been proposed to achieve better distillation performance~\cite{fitnets,at,ofd,crd}.
The working mechanism of knowledge distillation also attracts research attention~\cite{understandingkd1,understandingkd2,understandingkdji,understandingkd3,understandingkd4,understandingkd5,understanding0,tfkd}.
Yet, the optimization property of knowledge distillation has not been widely investigated, which is also an important perspective to understand KD.

As shown in Figure~\ref{fig:fig1}~(\textit{left}), the typical optimization objective of knowledge distillation is composed of two parts, a task loss~(\emph{e.g.}, the cross-entropy loss) and a distillation loss~(\emph{e.g.}, the KL-Divergence~\cite{kd}). We mainly study how the incremental distillation loss influences the optimization of task loss. 
Concretely, for an image classification task, we visualize (1) the task loss landscapes and (2) task and distillation loss dynamics during the optimization. 
As Figure~\ref{fig:fig1}~(\textit{middle}) illustrates, we observe that the distillation loss helps the student converge to flat minima, where the student tends to generalize better due to the robustness of flatter minima~\cite{li2018visualizing,minima1,minima4}.
However, as illustrated in Figure~\ref{fig:fig1}~(\textit{right}), introducing distillation loss brings about a \textbf{trade-off}. The task loss is not converged as sufficiently as the cross-entropy baseline, although the student's logits become similar to the teacher's.

We suppose that the trade-off is somehow \textit{counter-intuitive}. The reason is presented below: the teacher always yields a lower task loss than the student due to the larger model capacity. If the distillation loss is sufficiently optimized, the task loss would also be decreased since the student becomes more similar to the better-performing teacher.
We ask: \emph{why is there a trade-off and how to break it?} We attempt to answer this question from the following perspective.
The task and distillation loss terms are combined with a simple summation in practical implementations of popular distillation methods~\cite{kd,fitnets,at,crd,ofd}. It could make the optimization manner degrade to multi-task learning, where the network attempts to find a balance between the two tasks. As aforementioned, if the distillation loss is sufficiently optimized, \textit{both} losses would be decreased. Thus, we believe that \textit{sufficiently optimizing the distillation loss is the key to breaking the trade-off}.

To this end, we present the Distillation-Oriented Trainer~(DOT) which enables the distillation loss to \textit{dominate} the optimization. It separately considers the gradients provided by the task and the distillation losses, then adjusts the optimization orientation by weighing different \emph{momentums}. A larger momentum is applied to distillation loss, while a smaller one is to task loss. It ensures the optimization is dominated by gradients of distillation loss since a larger momentum accumulates larger and more consistent gradients than a smaller one. In this way, DOT ensures a sufficient optimization of distillation loss.
We validate the effectiveness of DOT from three perspectives. (1) As illustrated in Figure~\ref{fig:fig1}~(\emph{right}), we prove that DOT successfully breaks the trade-off between task and distillation losses. (2) As illustrated in Figure~\ref{fig:fig1}~(\emph{middle}), our DOT achieves more generalizable and flatter minima, empirically supporting the benefits of DOT. (3) DOT improves performances of popular distillation methods without bells and whistles, achieving new SOTA results.

More importantly, our research brings new insights into the knowledge distillation community.
We show great potential for a better optimization manner of knowledge distillation. To the best of our knowledge, we provide the first attempt to understand the working mechanism of knowledge distillation from the optimization perspective.

\section{Related Work}

\vspace{4pt}\noindent\textbf{Knowledge distillation.} Ideas correlated to distillation can date back to~\cite{dateback1,dateback2,dateback3,dateback4}, and the knowledge distillation concept has become widely known since the application in compressing a heavy network into a light one~\cite{kd}. Following representative works can be divided into two categories, \emph{i.e.}, distilling knowledge from logits~\cite{kd,eskd,zhao2022decoupled,ban} and intermediate features~\cite{fitnets,at,rkd,ab,ofd,crd}. More and more attempts have been made on understanding how and why knowledge distillation helps the network learning~\cite{understandingkd1,understandingkd2,understandingkdji,understandingkd3,understandingkd4,understandingkd5,tfkd,understanding0} recently. KD~\cite{kd} conjectures that the improvement comes from network predictions on incorrect classes. \cite{understandingkd2} explains distillation from a privileged information perspective. \cite{understandingkd1} studies distillation with linear classifiers and proposes that the success of distillation owes to data geometry, optimization bias, and strong monotonicity. \cite{understandingkdji} explains the practice of mixing task loss and distillation loss with the data inefficiency concept. Few efforts are made to analyze distillation from the network optimization perspective, while we provide the first attempt in this work.

\vspace{4pt}\noindent\textbf{Flatness of minima.} The minima of neural networks has attracted great research attention~\cite{minima1,minima2,minima3,minima4,asymvalley}. An acknowledged hypothesis is that the flatness of converged minima can influence the generalization ability of networks~\cite{minima1}, and flatter ones correspond to better generalization ability. The explanation is that flat ones reduce generalization errors since there are random perturbations around the loss landscape and the flatter one is more robust. \cite{asymvalley} also proves the correctness of this hypothesis. Thanks to previous efforts in analyzing loss landscapes~\cite{li2018visualizing,visualization1,visualization2,visualization3}, we are allowed to visualize the converged minima learned by knowledge distillation to understand its working mechanism.

In this work, we investigate the optimization process of the most representative method KD by visualizing the flatness of minima for the first time. We show that introducing distillation loss benefits the student with better generalization flat minima while resulting in higher task loss. This trade-off between the task and distillation losses is counter-intuitive and mostly overlooked. To this end, we study the optimization process and propose a method to break the trade-off and approach ideal converged minima.

\section{Revisiting Knowledge Distillation: An Optimization Perspective}
\label{sec:sec3}
\subsection{Recap of Knowledge Distillation}
The working mechanism of knowledge distillation methods has been explored from various perspectives~\cite{understandingkd1,understandingkd2,understandingkdji,understandingkd3,understandingkd4,understandingkd5,tfkd,understanding0}.
Previous works seldom delve into the optimization property of knowledge distillation. In this section, we explore the optimization behavior of knowledge distillation by studying how the incremental distillation loss influences the optimization property.

We study the most representative knowledge distillation method KD~\cite{kd} for easy understanding. The practical loss function of KD could be written as: 

\begin{equation}
	\label{eq:kd}
	\begin{aligned}
	\mathcal{L} = & \alpha \mathcal{L}_{\text{CE}}(x,y; \boldsymbol{\theta}) + (1 - \alpha) \mathcal{L}_{\text{KD}}(x, \boldsymbol{\phi}; \boldsymbol{\theta}) \\ = & \alpha H(y, \sigma(p)) + (1 - \alpha) D_{\text{KL }}(\sigma(p/T)\parallel \sigma(q/T)),
	\end{aligned}
\end{equation}
where the input and its label are denoted as $x$ and $y$. $\boldsymbol{\theta}$ and $\boldsymbol{\phi}$ are the parameters of the student and teacher networks. $p$ and $q$ are the output logits from student and teacher networks respectively. $H$ is the cross-entropy loss function, $D_{\text{KL}}$ is the KL-divergence, and $\sigma$ is the softmax function. The temperature $T$ is introduced to soften predictions and arise attention on negative logits.

Eqn.~(\ref{eq:kd}) manifests that the practical optimization objective $\mathcal{L}$ is composed of a task loss $\mathcal{L}_{\text{CE}}$ and a distillation loss $\mathcal{L}_{\text{KD}}$. We mainly study how the incremental $\mathcal{L}_{\text{KD}}$ influences the optimization property.
First, we explore the influence of distillation loss on loss landscapes which can be used to measure the generalization ability of the converged minima.
Second, we visualize the loss dynamics and reveal a \textit{trade-off} between task and distillation losses along the entire optimization process.

\subsection{Loss Landscapes}
\label{subsec:landscape}

\begin{figure}[th]
	\centering
    \includegraphics[width=.475\textwidth]{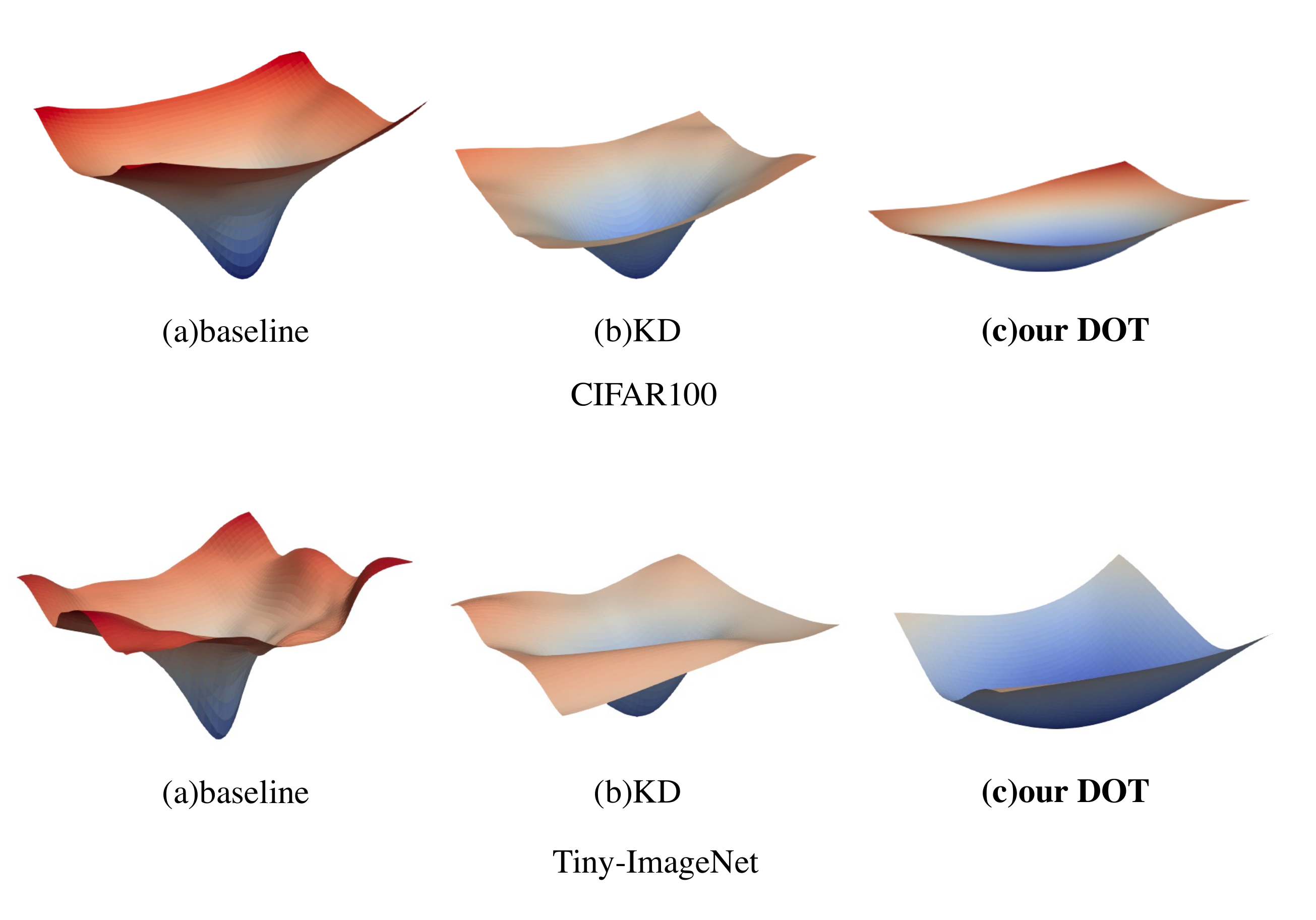}
	\caption{Loss landscapes on CIFAR-100 and Tiny-ImageNet. KD leads the student to flat minima with robust generalization ability. Our DOT achieves much \textit{flatter} minima, further improving the model generalization and distillation performances.}
\label{fig:fig2}
\end{figure}

It is well-acknowledged that the model generalization performance is characterized by the flatness of the local minima~\cite{minima1,minima2,minima3,minima4,asymvalley}. 
As illustrated in Figure~\ref{fig:fig1}~(\emph{middle}), sharp minima correspond to a large gap between training and test loss values, \ie, inferior generalization ability, but flat minima tend to reduce the generalization errors. After respectively training student networks with $\mathcal{L}_{\text{CE}}$~(the baseline) and $\mathcal{L}$~(the KD~\cite{kd}) on CIFAR-100~\cite{cifar}(typical ResNet32$\times$4-ResNet8$\times$4 pair) and Tiny-ImageNet~(typical ResNet18-MobileNetV2 pair), we visualize the task loss landscapes of the converged student networks to study the flatness of local minima in Figure~\ref{fig:fig2}. Compared to the baseline trained with only $\mathcal{L}_{\text{CE}}$, optimizing with $\mathcal{L}$ helps the task loss converge to flatter minima, which explains the generalization improvement of student networks. Conversely, training with only the task loss $\mathcal{L}_{\text{CE}}$ leads to the sharp minima, resulting in unsatisfactory generalization performance on the test distribution.
It suggests that the improvements made by knowledge distillation methods are attributed to enabling the student to converge around flatter minima.

\subsection{Trade-off Between Distillation and Task Losses}

\begin{figure}[th]
	\centering
	\subfloat[$\mathcal{L}_{\text{CE}}$]{
	    \includegraphics[width=.23\textwidth]{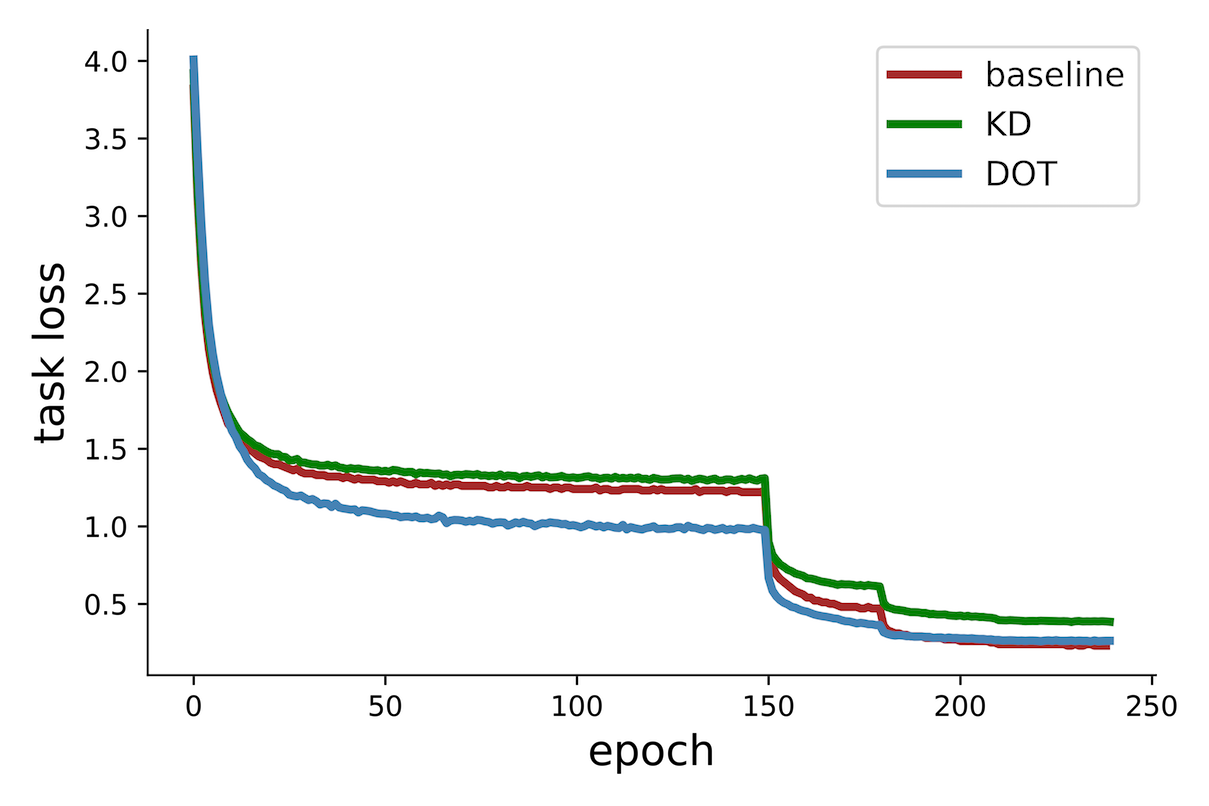}
	}
	\subfloat[$\mathcal{L}_{\text{KD}}$]{
	    \includegraphics[width=.23\textwidth]{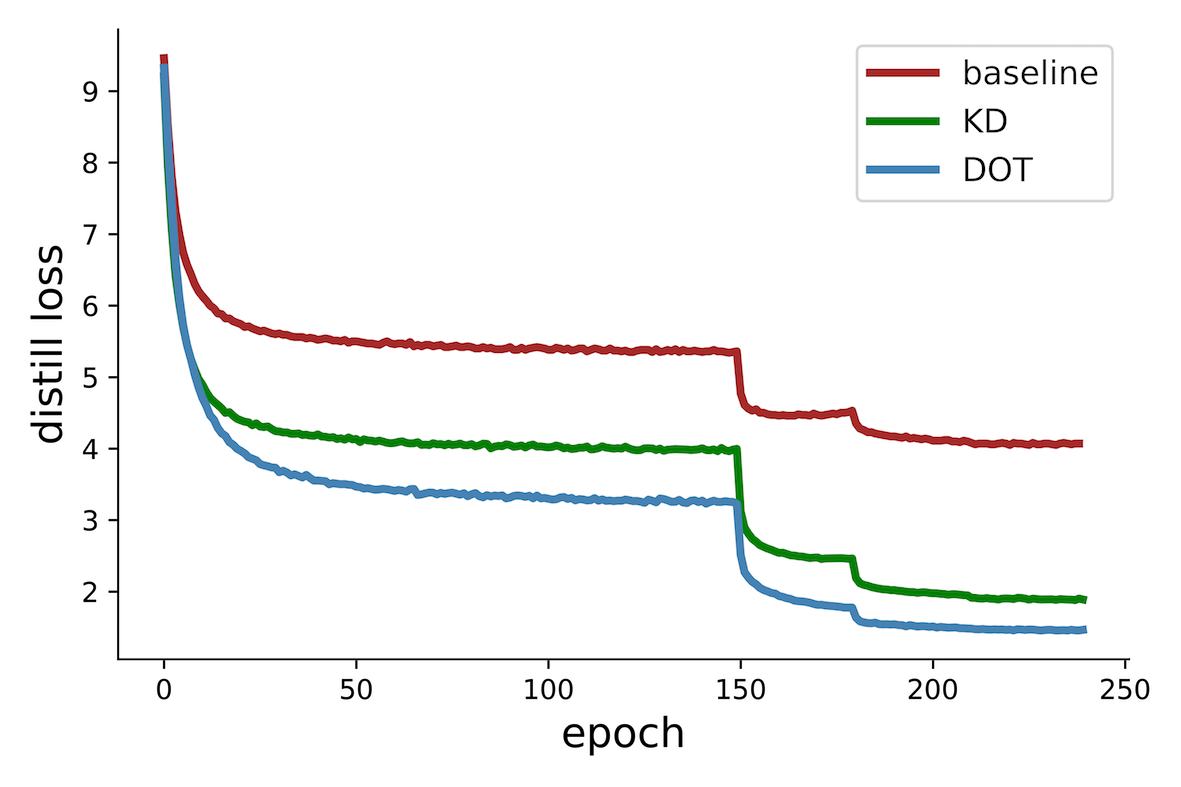}
	}
	\qquad
	\subfloat[$\mathcal{L}_{\text{CE}}$]{
	    \includegraphics[width=.23\textwidth]{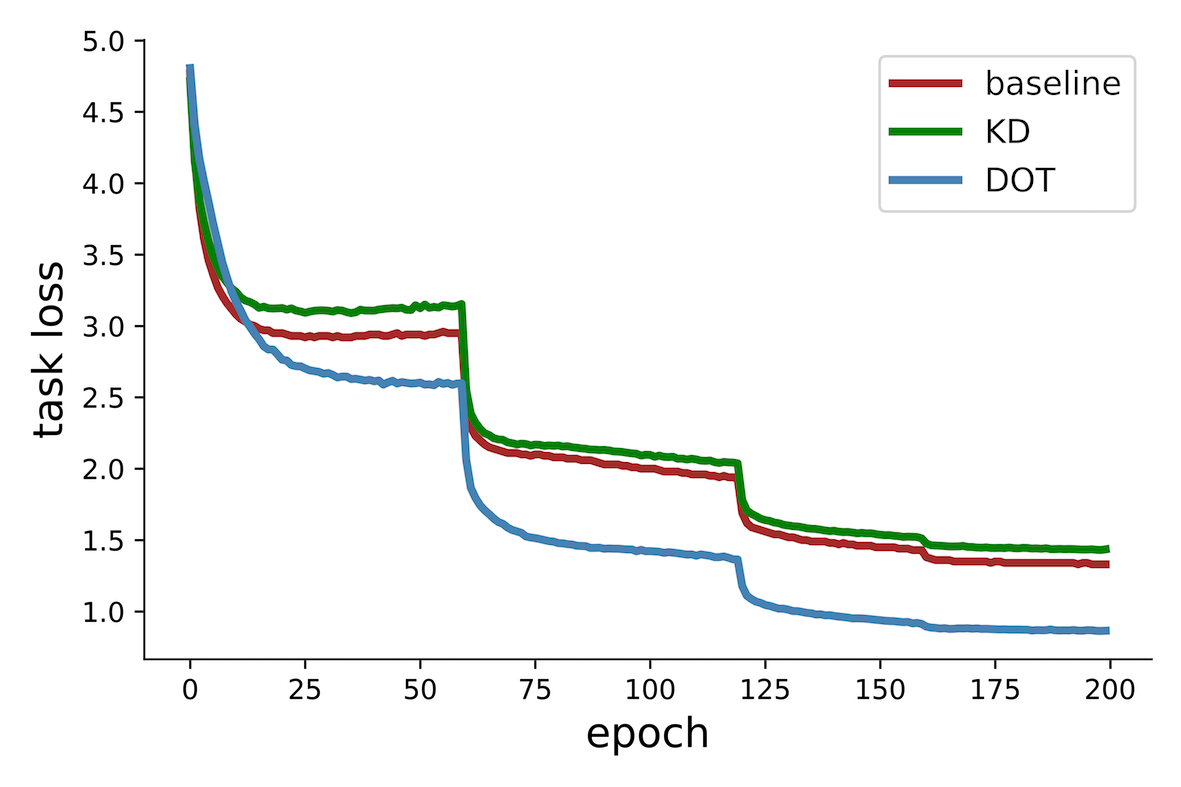}
	}
	\subfloat[$\mathcal{L}_{\text{KD}}$]{
	    \includegraphics[width=.23\textwidth]{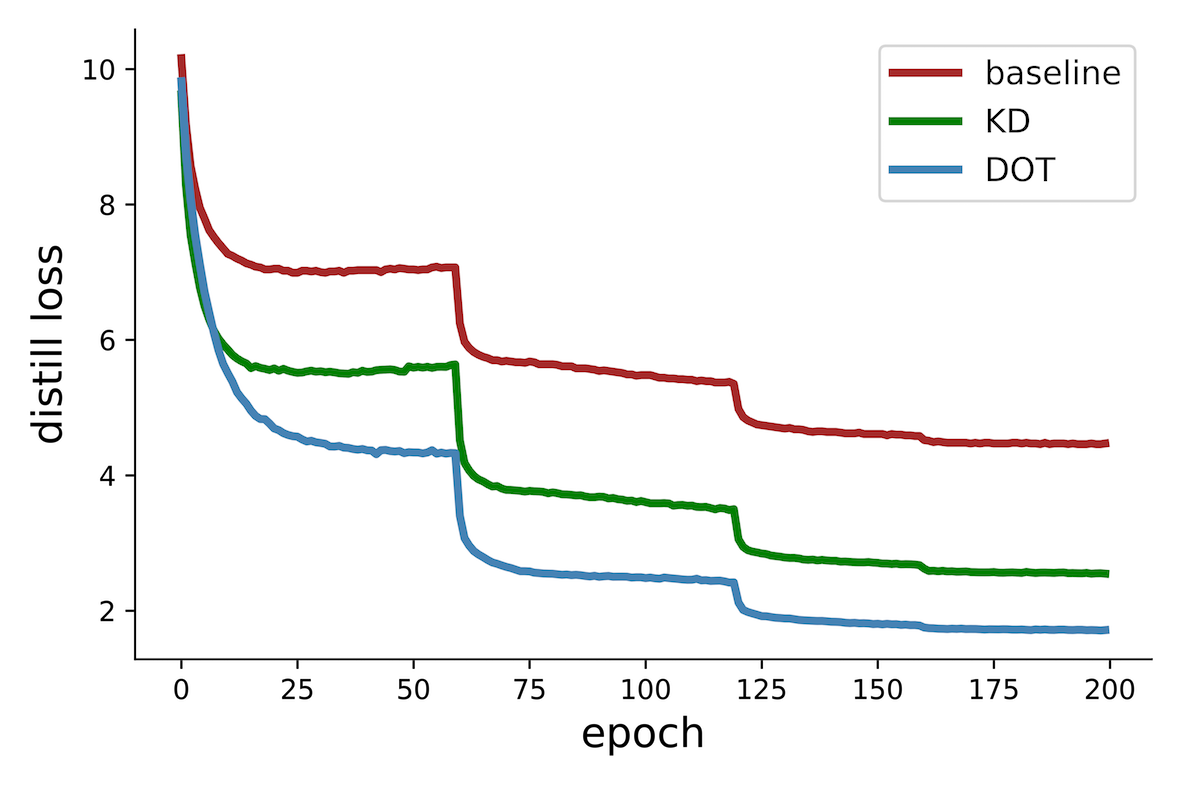}
	}
	\caption{Task and distillation loss curves on CIFAR-100 and Tiny-ImageNet. It suggests that there is a trade-off between task and distillation losses. Our proposed DOT could break the trade-off, achieving lower task and lower distillation losses \textit{at the same time}.}
\label{fig:fig2c}
\vspace{-10pt}
\end{figure}

In Figure~\ref{fig:fig2c}, we visualize the \textit{training} task and distillation loss dynamics with the progress of optimization. For the KD method, we respectively calculate the task loss and the distillation loss of each epoch. For the baseline, we are only allowed to obtain task loss, so we calculate the KL-Divergence between logits of the student and teacher on-the-fly (without back-propagation). The datasets and the teacher-student pairs are the same as those in Section~\ref{subsec:landscape}. As shown in Figure~\ref{fig:fig2c}, the introduction of distillation loss~(KD) greatly decreases $\mathcal{L}_{\text{KD}}$, because the output logits of the student become more similar to the teacher. However, the task loss $\mathcal{L}_{\text{CE}}$~(on training set) is increased. It indicates that the network attempts to find a trade-off between task and distillation losses. and the converged student could be a ``Pareto optimum''~\footnote{The trade-off can not be solved by simply tuning the loss weights, proofs are presented in Figure~\ref{fig:fig5} of Section~\ref{sec:sec5}}.

We attempt to explain the trade-off from the perspective of multi-task learning. The targets of both losses are not identical, and learning multiple tasks at the same time makes the optimization difficult~\cite{zhang2021survey}. Therefore, it is reasonable that a trade-off exists between task and distillation losses. And the aforementioned observations prove that the task training loss is increased due to the introduction of distillation loss~\footnote{We also conduct experiments with longer training time in the supplement, to validate that the high loss value is not due to inadequate training.}.
We suppose that regarding the optimization of task and distillation losses as multiple tasks is improper for the following reason: The teacher's training and test losses are both lower than the student's due to the larger model capacity. Making the student similar to the teacher can help yield \textit{both} lower distillation \textit{and} test losses. Thus, if the distillation loss can be sufficiently optimized, both task loss and distillation loss would be decreased. It inspires us to design an optimization manner where the distillation loss could be more sufficiently optimized.

\begin{figure*}[th]
	\centering
	\includegraphics[width=0.9\textwidth]{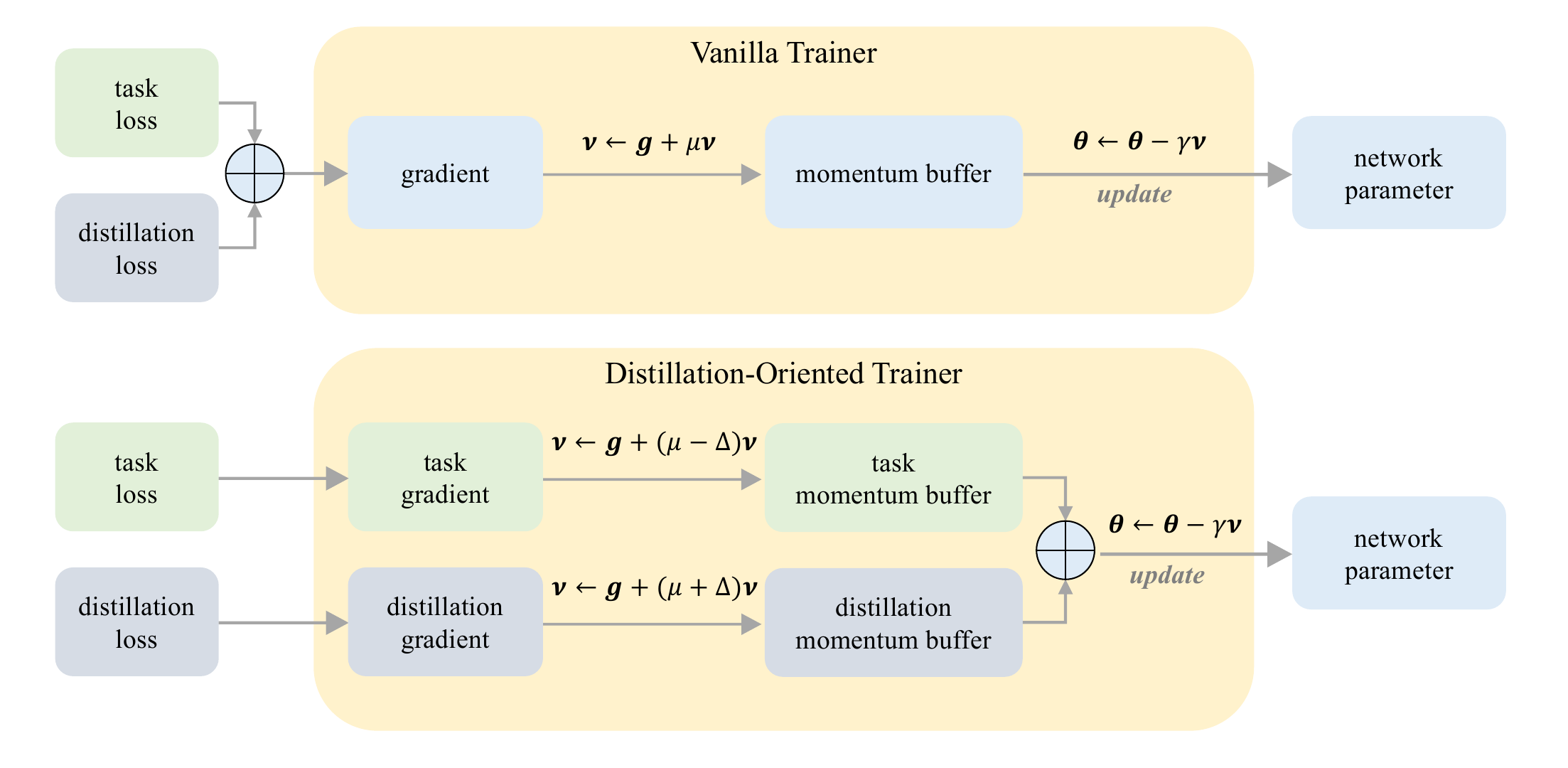}
	\vspace{-15pt}
	\caption{Illustration of a vanilla trainer and our Distillation-Oriented Trainer~(DOT). DOT \textit{separately} calculates the gradients of task and distillation losses, then applies \textit{larger momentum} to the distillation gradients and \textit{smaller momentum} to the task ones.}
\label{fig:method}
\end{figure*}

\section{Method}
\label{sec:sec4}
Knowledge distillation benefits the student network with flatter minima, yet introduces a trade-off between the task and the distillation losses. We suppose that the key to breaking the trade-off is making the optimization have a \textit{dominant orientation}, which could reduce gradient conflicts and ensure better network convergence. Since the distillation loss helps the student similar to the teacher, making it dominant~(instead of the task loss) could leverage the knowledge and achieve better generalization ability. 

\subsection{Making KD Dominate the Optimization}
\vspace{4pt}\noindent\textbf{Optimizer with momentum.} Firstly, we revisit the widely-used Stochastic Gradient Descent~(SGD) optimizer. SGD~(with momentum) updates the network parameters~(denoted as $\boldsymbol{\theta}$) with both the current gradients~(denoted as $\boldsymbol{g}=\nabla_{\boldsymbol{\theta}}\mathcal{L}(\boldsymbol{\theta})$) and the historical ones. Specifically, SGD maintains a grad buffer named ``momentum buffer''(denoted as $\boldsymbol{v}$) for network parameters. For every training mini-batch data, SGD updates the momentum buffer by: 
\begin{equation}
\boldsymbol{v} \leftarrow \boldsymbol{g} + \mu \boldsymbol{v},
\end{equation}
and then the parameters will be updated following the gradient-descent rule:
\begin{equation}
\boldsymbol{\theta} \leftarrow \boldsymbol{\theta} - \gamma \boldsymbol{v},
\end{equation}
where $\mu$ and $\gamma$ denote the momentum coefficient and the learning rate, respectively. Utilizing the momentum buffer can benefit the optimization process with the historical gradients. \cite{polyak1964some} shows that using momentum can considerably accelerate convergence to a local minimum. Empirically, the momentum coefficient $\mu$ is not the larger the better, and $0.9$ is the most used value.

\noindent\textbf{Independent momentums for distillation and task losses.} Momentum is a widely-used technique for accelerating gradient descent that accumulates a velocity vector in directions of persistent reduction in the objective across iterations~\cite{polyak1964some}. Under the knowledge distillation framework, setting independent momentums for the grads provided by different losses~(\emph{i.e.}, the distillation loss and the task loss) could play an important role in controlling the optimization orientation. Independent momentums enable the loss with the larger momentum to dominate the optimization from two aspects: (1) A large momentum on the distillation loss ensures the optimization orientation is knowledge-transfer-friendly in the initial ``transient phase''~\cite{darken1991towards}. (2) A large momentum keeps the historical gradient value undiminished in later training, ensuring the consistency of optimization orientation. 

\noindent\textbf{Distillation Oriented Trainer.} Driven by the analysis above, we present Distillation-Oriented Trainer~(DOT)~\footnote{DOT can be applied to all optimization methods with momentum mechanism. In this paper, we apply DOT to SGD since it is the most common one in the knowledge distillation community.}. As shown in Figure~\ref{fig:method}, DOT maintains two individual momentum buffers for the gradients of CE loss and KD loss. The two momentum buffers are denoted as $\boldsymbol{v}_{\text{ce}}$ and $\boldsymbol{v}_{\text{kd}}$. DOT updates $\boldsymbol{v}_{\text{ce}}$ and $\boldsymbol{v}_{\text{kd}}$ with different momentum coefficients. DOT introduces a hyper-parameter $\Delta$ and sets the coefficients for $\boldsymbol{v}_{\text{ce}}$ and $\boldsymbol{v}_{\text{kd}}$ as $\mu - \Delta$ and $\mu + \Delta$, respectively. Given a single mini-batch data, DOT first computes the gradients~(denoted as $\boldsymbol{g}_{\text{ce}}$ and $\boldsymbol{g}_{\text{kd}}$) produced by $\mathcal{L}_{\text{CE}}$ and $\mathcal{L}_{\text{KD}}$ respectively, then updates the momentum buffers according to: 
\begin{equation}
\begin{aligned}
&\boldsymbol{v}_{\text{ce}}\leftarrow \boldsymbol{g}_{\text{ce}} + (\mu - \Delta) \boldsymbol{v}_{\text{ce}}, \\
&\boldsymbol{v}_{\text{kd}}\leftarrow \boldsymbol{g}_{\text{kd}} + (\mu + \Delta) \boldsymbol{v}_{\text{kd}}.
\end{aligned}
\end{equation}
Finally, the network parameters are updated with the sum of two momentum buffers:
\begin{equation}
\boldsymbol{\theta} \leftarrow \boldsymbol{\theta} - \gamma (\boldsymbol{v}_{\text{ce}} + \boldsymbol{v}_{\text{kd}})
\end{equation}

DOT applies larger momentum to the distillation loss $\mathcal{L}_{\text{KD}}$ and smaller momentum to the task loss $\mathcal{L}_{\text{CE}}$. Thus, the optimization orientation could be dominated by the gradients of the distillation loss. DOT better leverages the knowledge from the teacher and mitigates the trade-off problem caused by insufficient optmization~\footnote{The algorithm of DOT and more details about DOT's implementation are attached in the supplement.}.

\subsection{Theoretical Analysis of Gradient}

To investigate the difference between DOT and a vanilla trainer, we dissect the gradients of task and distillation losses as:
\begin{equation}
    \begin{aligned}
    &\boldsymbol{v}_{\text{ce}} = \boldsymbol{v}^{\text{con}} + \boldsymbol{v}_{\text{ce}}^{\text{incon}}, \\
    &\boldsymbol{v}_{\text{kd}} = \boldsymbol{v}^{\text{con}} + \boldsymbol{v}_{\text{kd}}^{\text{incon}}.
    \end{aligned}
\end{equation}
$\boldsymbol{v}^{\text{con}}$ denotes the ``consistent'' gradient component of both losses. $\boldsymbol{v}_{\text{ce}}^{\text{incon}}$ and $\boldsymbol{v}_{\text{kd}}^{\text{incon}}$ are the ``inconsistent'' components.
For the SGD baseline, the updated momentum buffer can be written as :
\begin{equation}
    \begin{aligned}
    \boldsymbol{v}_{\text{sgd}} &= \boldsymbol{g}_{\text{ce}} + \boldsymbol{g}_{\text{kd}} + \mu (\boldsymbol{v}_{\text{ce}} + \boldsymbol{v}_{\text{kd}}) \\
    &= \boldsymbol{g}_{\text{ce}} + \boldsymbol{g}_{\text{kd}} + \mu (\boldsymbol{v}_{\text{ce}}^{\text{incon}} + \boldsymbol{v}_{\text{kd}}^{\text{incon}} + 2\boldsymbol{v}^{\text{con}})
    \end{aligned}
\label{eq:eq_sgd}
\end{equation}
For DOT, the updated momentum buffer is :
\begin{equation}
    \begin{aligned}
    \boldsymbol{v}_{\text{dot}} &= \boldsymbol{g}_{\text{ce}} + \boldsymbol{g}_{\text{kd}} + (\mu - \Delta)\boldsymbol{v}_{\text{ce}} + (\mu + \Delta)\boldsymbol{v}_{\text{kd}}\\
    &=\boldsymbol{g}_{\text{ce}} + \boldsymbol{g}_{\text{kd}} + (\mu - \Delta)(\boldsymbol{v}_{\text{ce}}^{\text{incon}} + \boldsymbol{v}^{\text{con}}) \\& + (\mu + \Delta)(\boldsymbol{v}_{\text{kd}}^{\text{incon}} + \boldsymbol{v}^{\text{con}})
    \end{aligned}
\label{eq:eq_dot}
\end{equation}
Comparing Eqn.~(\ref{eq:eq_sgd}) and Eqn.~(\ref{eq:eq_dot}), the difference between the two methods is calculated as follows:
\begin{equation}
\label{eq:v_diff}
    \begin{aligned}
    \boldsymbol{v}_{\text{diff}} &= \boldsymbol{v}_{\text{dot}} - \boldsymbol{v}_{\text{sgd}} \\
    &= \Delta(\boldsymbol{v}_{\text{kd}}^{\text{incon}} - \boldsymbol{v}_{\text{ce}}^{\text{incon}}). 
    \end{aligned}
\end{equation}
Eqn.~(\ref{eq:v_diff}) indicates that when the gradient of task loss conflicts with that of distillation loss, DOT produces gradient $\Delta(\boldsymbol{v}_{\text{kd}}^{\text{incon}} - \boldsymbol{v}_{\text{ce}}^{\text{incon}})$ to accelerate the accumulation of the gradient of the distillation loss. Thus, the optimization is driven by the orientation of distillation loss~\footnote{Besides, we also conduct a toy experiment for better understanding our DOT in the supplement.}. The student network would achieve better convergence since the conflict between gradients is alleviated.

\section{Experiments}
\label{sec:sec5}

In this part, we first detail the implementations of all experiments. Second, we empirically validate our motivations and then discuss the main results on popular image classification benchmarks. We also provide analysis~(\emph{e.g.}, visualizations) for more insights.
 
\subsection{Experimental Settings}
\label{sec:sec5_impl}

\vspace{4pt}\noindent\textbf{Datasets.} We conduct experiments on three image classification datasets, CIFAR-100~\cite{cifar}, Tiny-ImageNet and ImageNet-1k~\cite{imagenet}. CIFAR-100 is a well-known image classification dataset that consists of 100 classes. The image size is $32\times32$. Training and validation sets are composed of 50k and 10k images, respectively. Tiny-ImageNet consists of 200 classes and the image size is $64\times64$. The training set contains 100k images and the validation contains 10k images. ImageNet-1k is a large-scale classification dataset that consists of 1k classes. The training set contains 1.28 million images and the validation contains 50k images. All images are cropped to $224\times224$.

\vspace{4pt}\noindent\textbf{Implementations.} All models are trained three times and we report the average accuracy.

\emph{For CIFAR-100}, we follow the settings in \cite{crd}. We train all models for 240 epochs with learning rates decayed by 0.1 at 150th, 180th, and 210th epoch. The initial learning rate is set as 0.01 for MobileNetV2~\cite{mobilenetv2} and ShuffleNetV2~\cite{shufflenetv2}, and 0.05 for ResNet$8\times4$. The batch size is 64 for all models. We use SGD with 0.9 momentum and 0.0005 weight decay as the optimizer. We find $\Delta$ ranging from 0.05 to 0.075 works well in all experiments. All models are trained on a single GPU. 

\emph{For Tiny-ImageNet}, we follow the settings in \cite{tfkd}. All the models are trained for 200 epochs with learning rates decayed by 0.1 at 60th, 120th, and 160th epoch. The initial learning rate is 0.05 for a 64 batch size. We use SGD with 0.9 momentum and 0.0005 weight decay as the optimizer. The hyper-parameter $\Delta$ for our DOT is set to 0.075. All models are trained on 4 GPUs.

\emph{For ImageNet-1k}, we use the standard training recipe in \cite{crd}. All models are trained for 100 epochs and the learning rates decayed by 0.1 at 30th, 60th, 90th epoch. The initial learning rate is 0.2 for a 512 batch size. We use SGD with 0.9 momentum and 0.0001 weight decay as the optimizer. The hyper-parameter $\Delta$ for our DOT is set to 0.09. All models are trained on 8 GPUs.

\subsection{Motivation Validations}

The verification of our conjectures and motivations is the most important experiment. In this part, We mainly validate that (1) the distillation loss is the dominant component, (2) the trade-off problem is alleviated, and (3) the student is benefited with reasonable flatter minima compared with the classical KD. Additionally, we also conduct ablation experiments for $\mu$ and $\Delta$.

\begin{figure}[ht]
	\centering
    \subfloat[\small{$\text{cos}(v_{kd}, v)$}]{\includegraphics[width=0.235\textwidth]{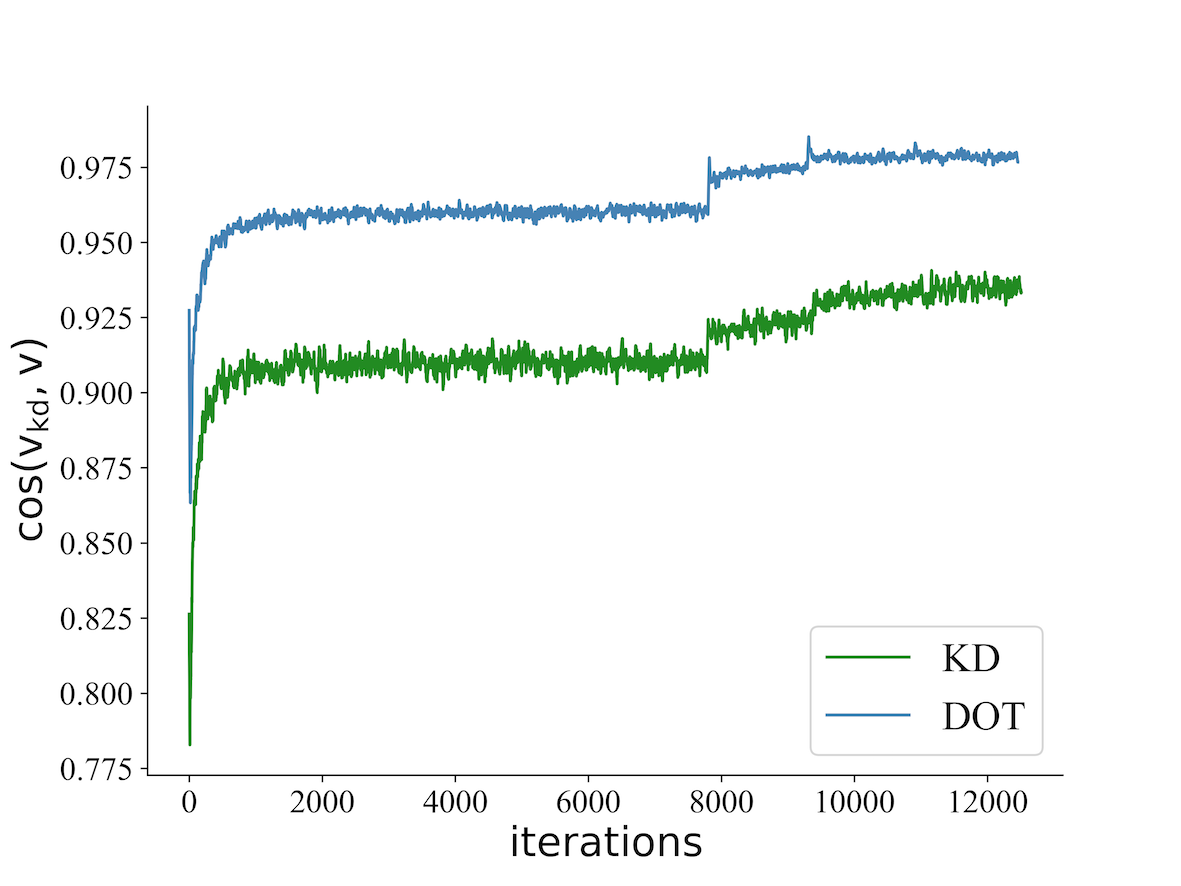}}
	\subfloat[\small{$\text{cos}(v_{ce}, v)$}]{\includegraphics[width=0.235\textwidth]{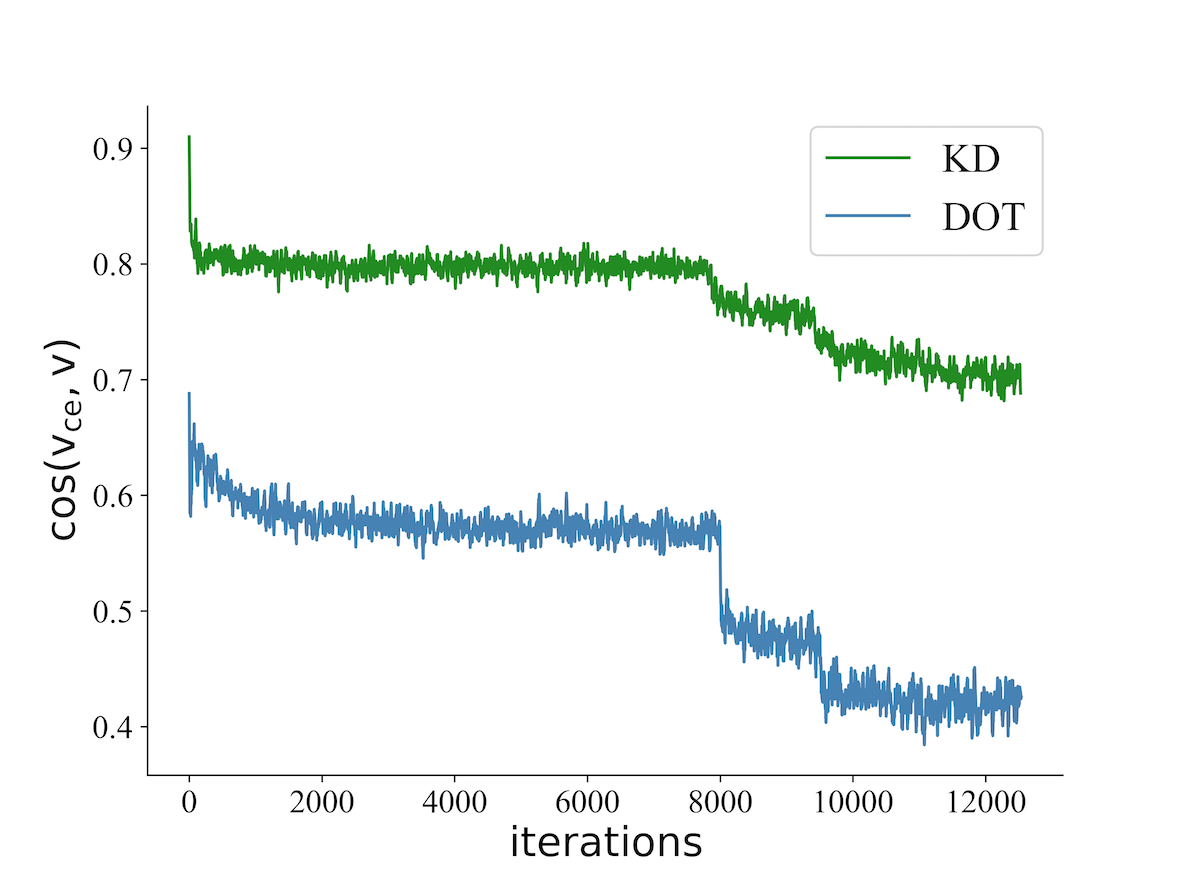}}
\caption{Cosine similarities. We visualize the $\text{cos}(\boldsymbol{v}_{\text{kd}}, \boldsymbol{v})$ and $\text{cos}(\boldsymbol{v}_{\text{ce}}, \boldsymbol{v})$ of each training iteration. It indicates that applying DOT greatly increases $\text{cos}(\boldsymbol{v}_{\text{kd}}, \boldsymbol{v})$, showing the momentum gradients are \textit{dominated} by the distillation component.}
\label{fig:cos_vis}
\vspace{-10pt}
\end{figure}

\vspace{4pt}\noindent\textbf{Does KD Loss dominate the optimization?}
To verify that DOT makes KD loss dominate the optimization, we analyze the gradients of DOT and a vanilla trainer in the optimization process. As shown in Figure~\ref{fig:cos_vis}, DOT leads to the following consequences: (1)~The cosine similarity between gradients of distillation and total losses is significantly increased, and (2)~cosine similarity between gradients of task and total losses is decreased. It suggests that the optimization orientation prefers distillation gradients after applying our DOT, which proves that DOT enables the distillation loss to dominate the optimization.

\vspace{4pt}\noindent\textbf{Does KD and Task Losses converge better?}
We visualize the training loss curves to study the convergence of task and distillation losses. As shown in Figure~\ref{fig:fig2c}, DOT \textit{simultaneously} achieves lower task and distillation losses than the vanilla trainer and the cross-entropy baseline. It strongly supports our motivation that sufficiently optimized distillation loss contributes to low task loss. Moreover, it also demonstrates that dominating the optimization with distillation loss eliminates the trade-off between task and distillation losses.

\vspace{4pt}\noindent\textbf{Does DOT lead to better minima?} As shown in Figure~\ref{fig:fig2} we provide visualizations based on experimental settings in Section~\ref{sec:sec3}, which reveal the effect of applying DOT to the representative method KD~\cite{kd}. Concretely, the loss landscapes of baseline, KD, and DOT on the CIFAR-100 training set are respectively illustrated. It is proven that DOT leads the student to minima with \textit{great flatness}, which is even flatter than the ones of KD. Conclusively, DOT achieves the goal to learn minima with both good flatness and low task loss, as also illustrated in Figure~\ref{fig:fig1}~(\textit{middle}).

\begin{table}[bth]
\begin{small}
\centering
    \centering
    
    \begin{tabular}{cccccc}
        $\Delta$ & \textbf{0.0} & +0.025 & +0.05 & +0.075 & +0.09 \\ \Xhline{3\arrayrulewidth}
        top-1 & 73.33 & 74.08 & 74.22 & \textbf{75.12} & 74.43 \\
    \end{tabular}
    \qquad
    
    \vspace{3pt}
    \centering
    \begin{tabular}{cccccc}
        $\Delta$ & -0.09 & -0.075 & -0.05 & -0.025 & \textbf{0.0} \\ \Xhline{3\arrayrulewidth}
        top-1 & 61.58 & 68.88 & 72.77 & 72.86 & 73.33 \\
    \end{tabular}
\caption{Different $\Delta$ on CIFAR-100. The results indicate that ``distillation-oriented"~(positive $\Delta$) could achieve \textbf{stable} improvements among different hyper-parameter values while ``task-oriented"~(negative $\Delta$) leads to performance drops.}
\label{tab:abl1}
\end{small}
\vspace{-10pt}
\end{table}

\begin{table}[bth]
\begin{small}
\centering
    \centering
    \begin{tabular}{cccccc}
        $\mu$ & 0.81 & 0.825 & 0.85 & 0.875 & \textbf{0.9} \\ \Xhline{3\arrayrulewidth}
        top-1 &  73.53 & 73.66 & 73.45 & 73.21 & 73.33
    \end{tabular}
    \qquad
    \vspace{3pt}
    \centering
    \begin{tabular}{cccccc}
        $\mu$ & \textbf{0.9} & 0.925 & 0.95 & 0.975 & 0.99  \\ \Xhline{3\arrayrulewidth}
        top-1 & 73.33 & 73.16 & 73.14 & 68.75 & 37.19
    \end{tabular}
\caption{Ablation study for different $\mu$ on CIFAR-100. It indicates that tuning $\mu$ brings no significant performance improvement.}
\label{tab:abl2}
\end{small}
\vspace{-10pt}
\end{table}

\begin{table*}[th]
\centering
\begin{small}
\begin{subtable}{1.0\linewidth}
\centering
\begin{tabular}{ccc|cccc|ccc}
& teacher & student & KD & AT & CRD & DKD & KD+DOT & CRD+DOT & 
DKD+DOT \\
\Xhline{3\arrayrulewidth}
\multicolumn{10}{c}{\textit{ResNet32$\times$4 as the teacher, ResNet8$\times$4 as the student}} \\ \hline
top1 & 79.42 & 72.50 & 73.33 & 73.44 & 75.51 & 76.32 & \textbf{75.12}~\textcolor{PineGreen}{(+1.79)} & \textbf{75.99}~\textcolor{PineGreen}{(+0.48)} & \textbf{76.64}~\textcolor{PineGreen}{(+0.32)} \\
top5 & 94.58 & 92.72 &  93.00 & 93.06 & 93.92 & 93.94 & \textbf{93.46}~\textcolor{PineGreen}{(+0.46)} & \textbf{94.09}~\textcolor{PineGreen}{(+0.17)} & \textbf{94.03}~\textcolor{PineGreen}{(+0.09)} \\ \hline \hline
\multicolumn{10}{c}{\textit{VGG13 as the teacher, VGG8 as the student}} \\ \hline
top1 & 74.64 & 70.36 & 72.98 & 71.43 & 73.94 & 74.68 & \textbf{73.77}~\textcolor{PineGreen}{(+0.79)} & \textbf{74.21}~\textcolor{PineGreen}{(+0.27)} & \textbf{74.86}~\textcolor{PineGreen}{(+0.18)} \\
top5 & 92.60 & 90.57 & 92.27 & 91.94 & 92.25 & 92.62 & \textbf{92.33}~\textcolor{PineGreen}{(+0.05)} & \textbf{92.72}~\textcolor{PineGreen}{(+0.47)} & \textbf{92.81}~\textcolor{PineGreen}{(+0.19)} \\ \hline \hline
\multicolumn{10}{c}{\textit{ResNet32$\times$4 as the teacher, ShuffleNetV2 as the student}} \\ \hline
top1 & 79.42 & 71.82 & 74.45 & 72.73 & 75.65 & 77.07 & \textbf{75.55}~\textcolor{PineGreen}{(+1.10)} & \textbf{76.64}~\textcolor{PineGreen}{(+0.99)} & \textbf{77.41}~\textcolor{PineGreen}{(+0.34)} \\
top5 & 94.58 & 91.77 & 93.02 & 93.07 & 93.71 & 94.19 & \textbf{93.24}~\textcolor{PineGreen}{(+0.22)} & \textbf{94.00}~\textcolor{PineGreen}{(+0.29)} & \textbf{94.13}~\textcolor{PineGreen}{(-0.06)} \\ 
\end{tabular}
\label{tab:xifa}
\caption{CIFAR-100}
\end{subtable}

\vspace{5pt}
\begin{subtable}{1.0\linewidth}
\centering
\begin{tabular}{ccc|cccc|ccc}
& teacher & student & KD & AT & CRD & DKD & KD+DOT & CRD+DOT & 
DKD+DOT \\
\Xhline{3\arrayrulewidth}
\multicolumn{10}{c}{\textit{ResNet18 as the teacher, MobileNet-V2 as the student}} \\ \hline
top1 & 62.99 & 56.28 & 58.35 & 57.18 & 61.18 & 62.04 & \textbf{64.01}~\textcolor{PineGreen}{(+5.66)} & \textbf{64.12}~\textcolor{PineGreen}{(+2.94)} & \textbf{64.60}~\textcolor{PineGreen}{(+2.56)} \\
top5 & 83.36 & 80.32 & 82.07 & 81.52 & 83.13 & 84.12 &  \textbf{84.30}~\textcolor{PineGreen}{(+2.23)} & \textbf{84.43}~\textcolor{PineGreen}{(+1.30)} & \textbf{85.38}~\textcolor{PineGreen}{(+1.26)} \\ \hline \hline
\multicolumn{10}{c}{\textit{ResNet18 as the teacher, ShuffleNetV2 as the student}} \\ \hline
top1 & 62.99 & 60.78 & 62.26 & 62.45 & 63.97 & 65.06 & \textbf{65.75}~\textcolor{PineGreen}{(+3.49)} & \textbf{65.21}~\textcolor{PineGreen}{(+1.24)} & \textbf{66.21}~\textcolor{PineGreen}{(+1.15)} \\
top5 & 83.36 & 82.49 & 83.79 & 83.51 & 84.70 & 85.31 & \textbf{85.51}~\textcolor{PineGreen}{(+1.72)} & \textbf{85.13}~\textcolor{PineGreen}{(+0.43)} & \textbf{86.16}~\textcolor{PineGreen}{(+0.85)} \\
\end{tabular}
\label{tab:tiny}
\caption{Tiny ImageNet}
\end{subtable}

\vspace{5pt}
\begin{subtable}{1.0\linewidth}
\centering
\begin{tabular}{ccc|ccccc|cc}
& teacher & student & KD & AT & OFD & CRD & DKD & KD+DOT & DKD+DOT \\
\Xhline{3\arrayrulewidth}
\multicolumn{10}{c}{\textit{ResNet34 as the teacher, ResNet18 as the student}} \\ \hline
top1 & 73.31 & 69.75 & 71.03 & 70.69 & 70.81 & 71.17 & 71.70 & \textbf{71.72}~\textcolor{PineGreen}{(+0.69)} & \textbf{72.03}~\textcolor{PineGreen}{(+0.33)} \\
top5 & 91.42 & 89.07 & 90.05 & 90.01 & 89.98 & 90.13 & 90.05 & \textbf{90.30}~\textcolor{PineGreen}{(+0.25)} & \textbf{90.50}~\textcolor{PineGreen}{(+0.45)} \\ \hline \hline
\multicolumn{10}{c}{\textit{ResNet50 as the teacher, MobileNetV1 as the student}} \\ \hline
top1 & 76.16 & 68.87 & 70.50 & 69.56 & 71.25 & 71.37 & 72.05 & \textbf{73.09}~\textcolor{PineGreen}{(+2.59)} & \textbf{73.33}~\textcolor{PineGreen}{(+1.27)} \\
top5 & 92.86 & 88.76 & 89.80 & 89.33 & 90.34 & 90.41 & 91.05 & \textbf{91.11}~\textcolor{PineGreen}{(+1.31)} & \textbf{91.22}~\textcolor{PineGreen}{(+0.17)} \\
\end{tabular}
\label{tab:imgnet}
\caption{ImageNet-1k}
\end{subtable}
\end{small}

\caption{Applying our DOT can bring significant performance gains and achieve \textbf{new state-of-the-art} distillation results on CIFAR-100, Tiny-ImageNet and ImageNet-1k. Both logit- and feature-based methods could benefit from our DOT. Notably, for the ResNet50-MobileNetV1 pair on ImageNet-1k, DOT achieves a significant \textbf{+2.59\%} accuracy gain.}
\label{tab:main_result}
\end{table*}

\vspace{4pt}\noindent\textbf{Are independent momentums necessary?} To prove that the improvements come from the design of the independent momentum mechanism instead of carefully hyper-parameter tuning, we explore the influence of different hyper-parameter $\Delta$ for DOT. As shown in Table~\ref{tab:abl1}, setting $\Delta=0$ means training task and distillation losses with the same momentum~(\emph{i.e.}, equals to a vanilla SGD). Applying DOT with $\Delta>0$~(ranging from 0.025 to 0.09) enables the distillation loss to dominate optimization, which contributes to \textit{stable} performance improvements. It indicates that distillation-oriented optimization is of vital importance. Moreover, we also experiment with $\Delta<0$~(ranging from -0.09 to -0.025), corresponding to a ``task-oriented trainer'' where the task loss dominates the optimization. Significant performance drops could be observed in Table~\ref{tab:abl1}, which further supports our motivation to make the distillation loss dominant instead of the task loss.

\vspace{4pt}\noindent\textbf{Are improvements attributed to tuning momentum?} Without utilizing DOT, we also explore the distillation performance when training with different momentums $\mu$ to verify that the gains mainly come from our design of independent momentums instead of better momentum values. Results in Table~\ref{tab:abl2} show that carefully tuning $\mu$ could lead to certain performance gain~(73.33\% \textit{v.s.} 73.66\%), but not as significant as DOT's improvement~(73.33\% \textit{v.s.} 75.12\%).

\subsection{Main Results}

Following the common setting, we benchmark our DOT on three popular image classification datasets, \emph{i.e.}, CIFAR-100, Tiny-ImageNet, and ImageNet-1k. Additionally, we prove that DOT is compatible with representative distillation methods, and contributes to new state-of-the-art results.

\vspace{4pt}\noindent\textbf{CIFAR-100.} Results of CIFAR-100 in Table~\ref{tab:main_result}~(a) show that our DOT could contribute to significant performance gains for the classical knowledge distillation~\footnote{More pairs on CIFAR-100 can be attached to the supplement.}. For instance, DOT improves the classical KD method from 73.33\% to 75.12\% on the ResNet32$\times$4-ResNet8$\times$4 teacher-student pair. To prove the scalability of DOT, we combine DOT with popular logit-based and feature-based distillation methods. We select DKD~\cite{zhao2022decoupled} and CRD~\cite{crd} as representative logit-based and feature-based methods, respectively. As shown in Table~\ref{tab:main_result}~(a), DOT still succeeds in advancing the performances of evaluated methods, supporting DOT's practicability.

\vspace{4pt}\noindent\textbf{Tiny-ImageNet.} Tiny-ImageNet is a more challenging dataset than CIFAR-100. Results in Table~\ref{tab:main_result}~(b) demonstrate that our DOT still achieves \textit{more remarkable} performance gains on such a challenging dataset. DOT greatly improves the top1 accuracy from 58.35\% to 64.01\% on the ResNet18-MobileNetV2 teacher-student pair, and improves the top-1 accuracy from 62.26\% to 65.75\% on the ResNet18-ShuffleNetV2 pair. DOT also increases the SOTA performance to 64.60\% and 66.21\% on SOTA methods.

\vspace{4pt}\noindent\textbf{ImageNet-1k.} We also conduct experiments on ImageNet-1k. Experimental results in Table~\ref{tab:main_result}~(c) consistently validate the superiority of DOT. Especially for the ResNet50-MobileNetV1 pair, DOT achieves a +2.59\% accuracy gain on the classical KD method, outperforming previous state-of-the-art methods. It strongly demonstrates that the optimization of knowledge distillation methods deserves further exploration. Additionally, applying DOT to DKD could further increase the state-of-the-art performance to a new 73.33\% milestone.

\subsection{More Analysis}

In this part, we first investigate whether simply tuning the loss weight $\alpha$ can break the trade-off, then visualize the distillation fidelity for intuitive understanding.

\begin{figure}[ht]
\begin{tiny}
	\centering
     \subfloat[\small{$\mathcal{L}_{\text{CE}}$, CIFAR-100}]{\includegraphics[width=0.23\textwidth]{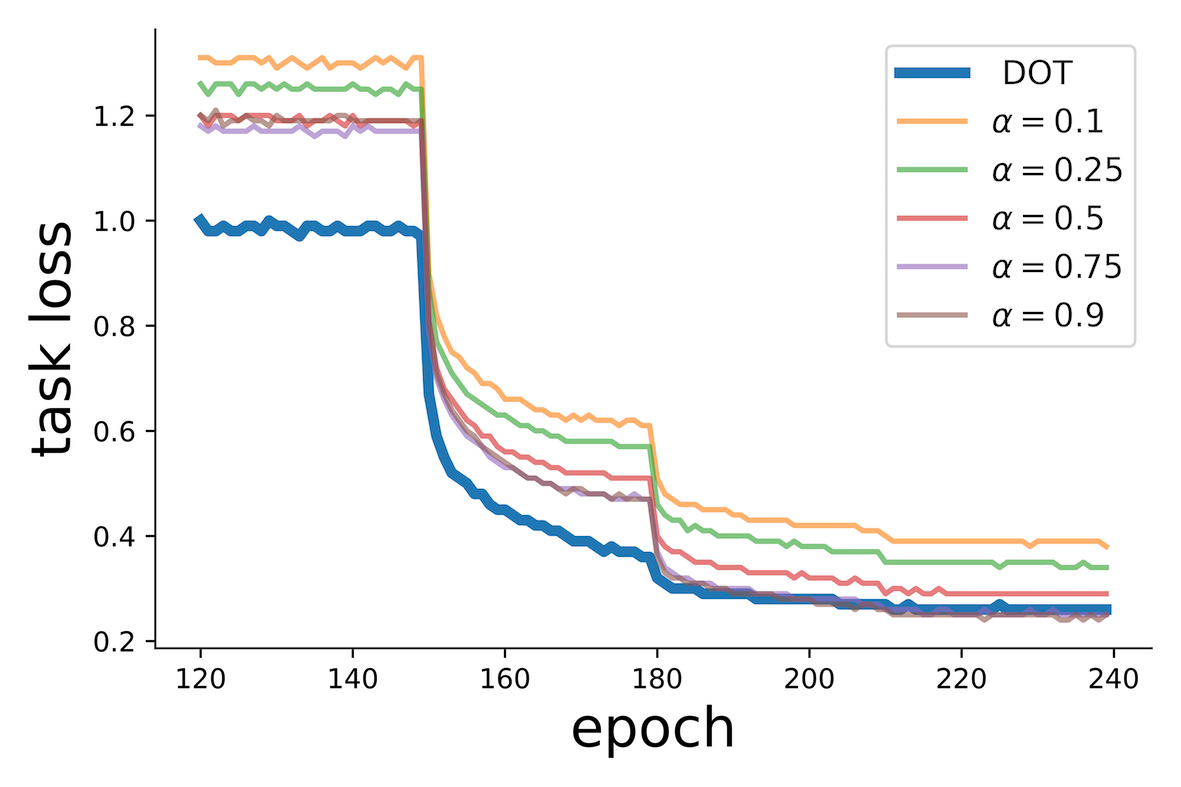}}
    \subfloat[\small{$\mathcal{L}_{\text{KD}}$, CIFAR-100}]{\includegraphics[width=0.23\textwidth]{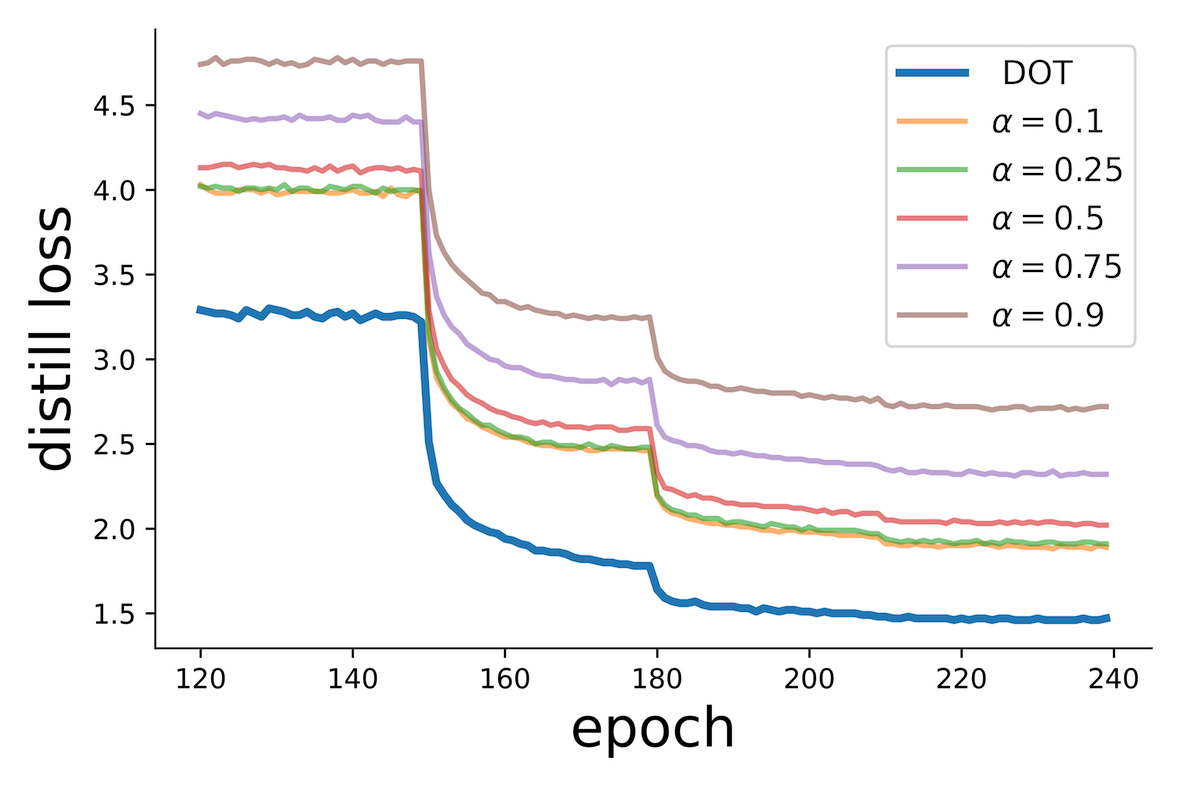}}
	\qquad
    \subfloat[\small{$\mathcal{L}_{\text{CE}}$, TinyImageNet}]{\includegraphics[width=0.23\textwidth]{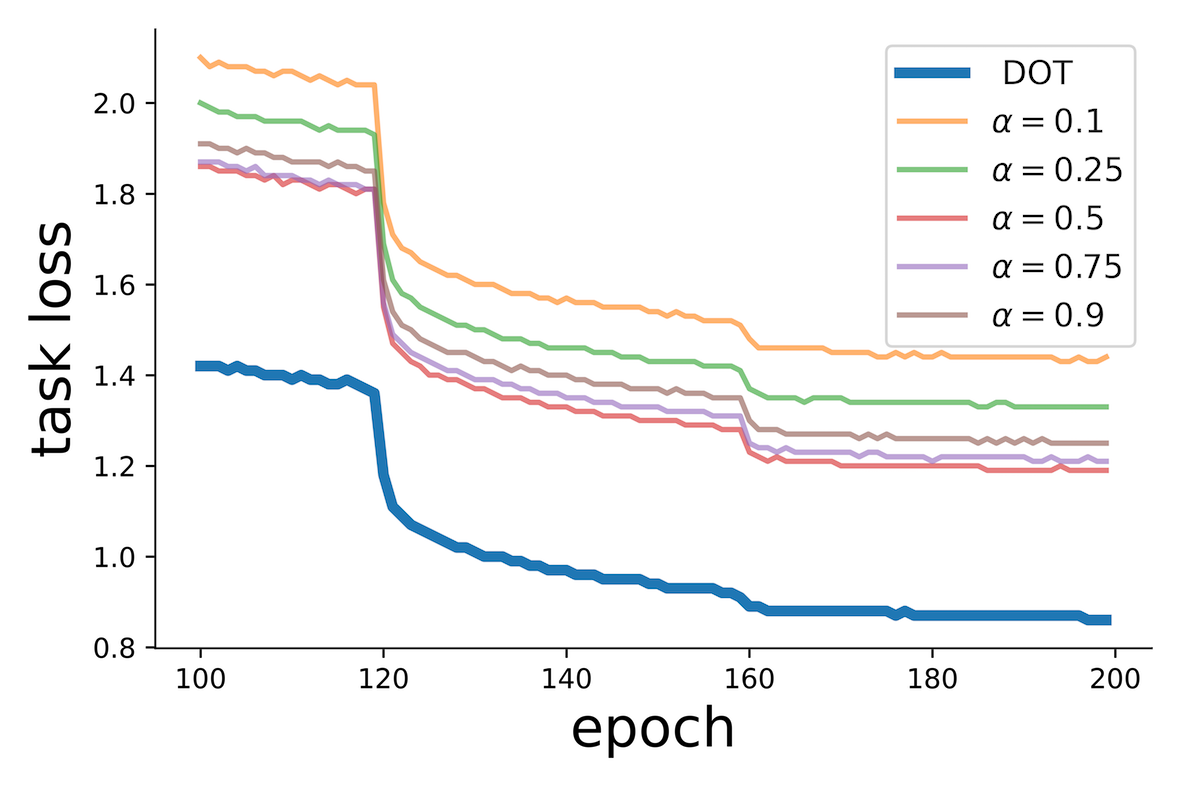}}
    \subfloat[\small{$\mathcal{L}_{\text{KD}}$, TinyImageNet}]{\includegraphics[width=0.23\textwidth]{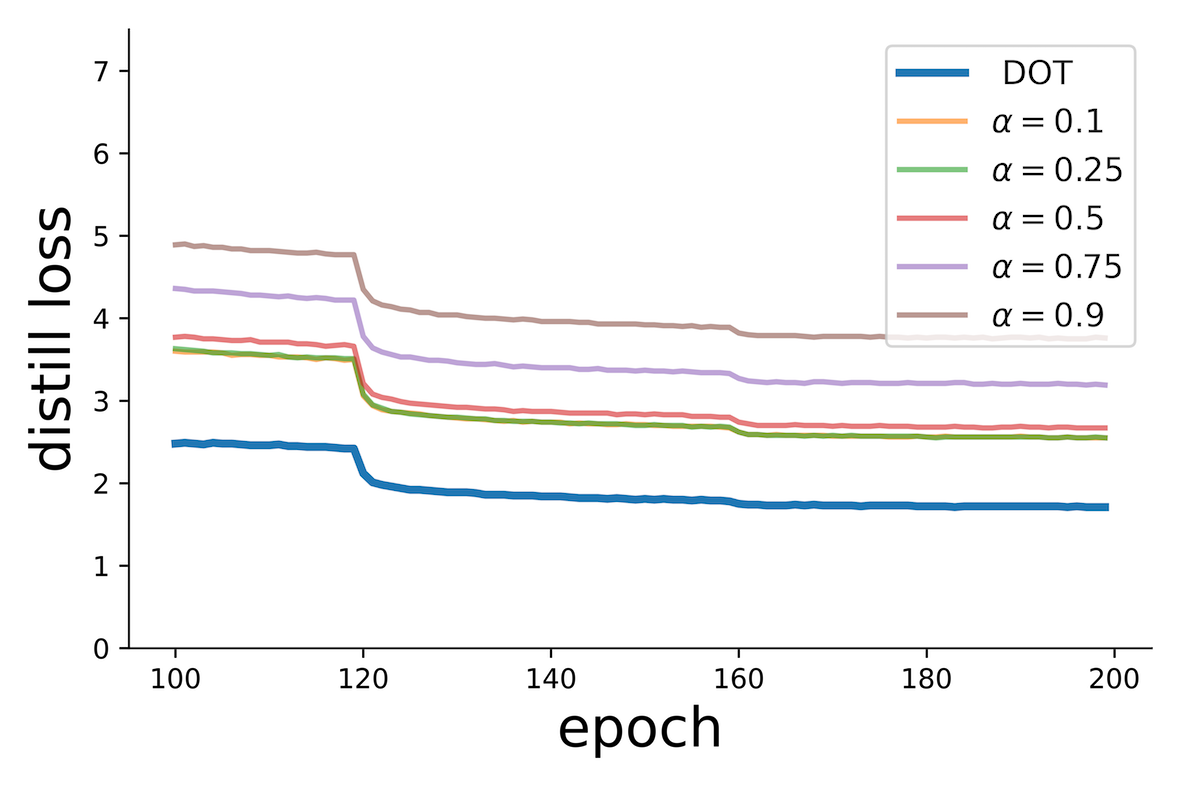}}
\end{tiny}
\caption{Illustration of loss curves for weighing $\alpha$. It indicates that simply tuning loss weights still suffers from the trade-off. A larger $\alpha$ leads to a lower task loss but a higher distillation loss.
In contrast, our DOT effectively \textbf{breaks the trade-off} and achieves both low task and distillation losses, \ie, better convergence.
}
\label{fig:fig5}
\vspace{-10pt}
\end{figure}

\vspace{4pt}\noindent\textbf{Adjusting $\alpha$ can not alleviate the trade-off.} Intuitively, adjusting $\alpha$ could somehow help the task loss converge better too. As shown in Table~\ref{tab:diff_weight}, Table~\ref{tab:diff_weight_tiny} and Figure~\ref{fig:fig5}, we adjust the weight $\alpha$ of the task loss~(and the weight of the distillation loss $1-\alpha$ is correspondingly adjusted) and provide the distillation results and the loss curves. Intuitively, the larger $\alpha$ strengthens the influence of the task loss on the optimization and weakens the influence of the distillation loss. Table~\ref{tab:diff_weight} and \ref{tab:diff_weight_tiny} show that weighing $\alpha$ can not exert noticeable influences on final distillation performances. What's more, Figure~\ref{fig:fig5} suggests that the larger $\alpha$ leads to lower task loss but higher distillation loss, which means the network still \textit{suffers from} the trade-off. Conversely, applying DOT achieves both lower task and distillation losses simultaneously, \ie, the trade-off between task and distillation losses is successfully alleviated.

\begin{table}[bth]
    \begin{minipage}{.45\textwidth}
    \centering
    \begin{small}
    \begin{tabular}{c|ccccc|c}
        $\alpha$ & 0.1 & 0.25 & 0.5 & 0.75 & 0.9 & \textbf{DOT} \\ \Xhline{3\arrayrulewidth}
        top-1 & 73.33 & 73.56 & 73.49 & 73.23 & 73.19 & \textbf{75.12} \\
    \end{tabular}
    \caption{Different weight~($\alpha$) for task loss on CIFAR-100. ResNet32$\times$4-ResNet8$\times$4 as the teacher-student pair.}
    \label{tab:diff_weight}
    \end{small}
    \end{minipage}
    \begin{minipage}{.45\textwidth}
    \centering
    \begin{small}
    \begin{tabular}{c|ccccc|c}
        $\alpha$ & 0.1 & 0.25 & 0.5 & 0.75 & 0.9 & \textbf{DOT} \\ \Xhline{3\arrayrulewidth}
        top-1 & 58.35 & 58.86 & 59.23 & 58.36 & 57.45 & \textbf{64.01}
    \end{tabular}
    \caption{Different weight~($\alpha$) for task loss on Tiny-ImageNet. ResNet18-MobileNetV2 as the teacher-student pair}
    \label{tab:diff_weight_tiny}
    \end{small}
    \end{minipage}
\vspace{-20pt}
\end{table}

\vspace{4pt}\noindent\textbf{Distillation fidelity. } We visualize the distillation fidelity following \cite{crd,zhao2022decoupled} for intuitive understanding. Concretely, for ResNet32$\times$4-ResNet8$\times$4~(CIFAR-100) and ResNet18-MobileNetV2~(Tiny-ImageNet) pairs, we calculate the absolute distance between correlation matrices of the teacher and the student. Compared with KD, introducing DOT helps the student output more similar logits to the teacher, resulting in better distillation performance.

\begin{figure}[ht]
\begin{tiny}
    \centering
    \begin{minipage}[t]{0.45\textwidth}
    \centering
	\subfloat[\small{KD, CIFAR-100}]{\includegraphics[width=0.4\textwidth]{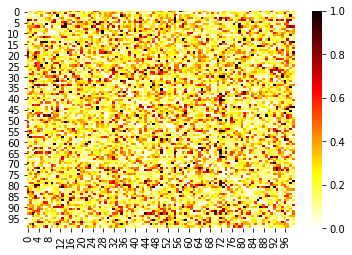}}
	\subfloat[\small{DOT, CIFAR-100}]{\includegraphics[width=0.4\textwidth]{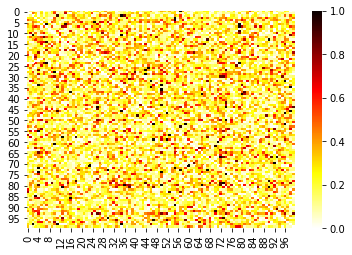}}
    \end{minipage}
    \qquad

    \begin{minipage}[t]{0.45\textwidth}
    \centering
	\subfloat[\small{KD, TinyImageNet}]{\includegraphics[width=0.4\textwidth]{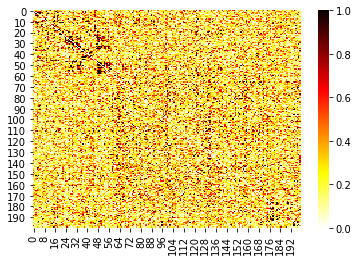}}
	\subfloat[\small{DOT, TinyImageNet}]{\includegraphics[width=0.4\textwidth]{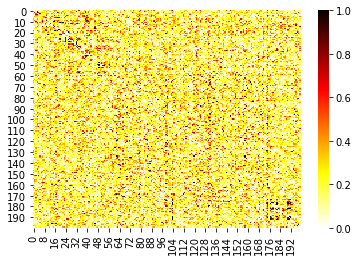}}
    \end{minipage}

\end{tiny}
\caption{Difference of student and teacher logits. DOT leads to a significantly smaller difference (more similar prediction) than KD.}
\label{fig:fig6}
\vspace{-10pt}
\end{figure}

\section{Conclusion}
In this paper, we investigate the optimization property of knowledge distillation. We reveal a counter-intuitive phenomenon that introducing distillation loss limits the convergence of task loss, \ie, a trade-off. We conjecture that the key to breaking the trade-off is sufficiently optimizing the distillation loss. To this end, we present a novel optimization method named Distillation-Oriented Trainer~(DOT). Extensive experiments validate our motivation and the practical value of DOT. Visualizations show that DOT leads to surprisingly flat minima with both lower task and distillation losses. Additionally, we demonstrate that DOT improves the performance of popular distillation methods. We hope this work can provide valuable experiences for future research in the knowledge distillation community.

{\small
\bibliographystyle{ieee_fullname}
\bibliography{egbib}

\begin{thebibliography}{10}\itemsep=-1pt

\bibitem{dateback1}
Cristian Buciluǎ, Rich Caruana, and Alexandru Niculescu-Mizil.
\newblock Model compression.
\newblock In {\em KDD}, 2006.

\bibitem{NIPS2017_e1e32e23}
Guobin Chen, Wongun Choi, Xiang Yu, Tony Han, and Manmohan Chandraker.
\newblock Learning efficient object detection models with knowledge
  distillation.
\newblock In {\em NeurIPS}, 2017.

\bibitem{eskd}
Jang~Hyun Cho and Bharath Hariharan.
\newblock On the efficacy of knowledge distillation.
\newblock In {\em ICCV}, 2019.

\bibitem{collobert2008unified}
Ronan Collobert and Jason Weston.
\newblock A unified architecture for natural language processing: Deep neural
  networks with multitask learning.
\newblock In {\em ICML}, 2008.

\bibitem{dateback2}
Mark Craven and Jude Shavlik.
\newblock Extracting tree-structured representations of trained networks.
\newblock In {\em NeurIPS}, 1995.

\bibitem{darken1991towards}
Christian Darken and John Moody.
\newblock Towards faster stochastic gradient search.
\newblock In {\em NeurIPS}, 1991.

\bibitem{devlin2018bert}
Jacob Devlin, Ming-Wei Chang, Kenton Lee, and Kristina Toutanova.
\newblock Bert: Pre-training of deep bidirectional transformers for language
  understanding.
\newblock {\em arXiv:1810.04805}, 2018.

\bibitem{minima4}
Laurent Dinh, Razvan Pascanu, Samy Bengio, and Yoshua Bengio.
\newblock Sharp minima can generalize for deep nets.
\newblock In {\em ICML}, 2017.

\bibitem{visualization3}
Felix Draxler, Kambis Veschgini, Manfred Salmhofer, and Fred Hamprecht.
\newblock Essentially no barriers in neural network energy landscape.
\newblock In {\em ICML}, 2018.

\bibitem{ban}
Tommaso Furlanello, Zachary Lipton, Michael Tschannen, Laurent Itti, and Anima
  Anandkumar.
\newblock Born again neural networks.
\newblock In {\em ICML}, 2018.

\bibitem{visualization1}
Ian~J Goodfellow, Oriol Vinyals, and Andrew~M Saxe.
\newblock Qualitatively characterizing neural network optimization problems.
\newblock {\em arXiv:1412.6544}, 2014.

\bibitem{survey2}
Jianping Gou, Baosheng Yu, Stephen~John Maybank, and Dacheng Tao.
\newblock Knowledge distillation: {A} survey.
\newblock {\em arXiv:2006.05525}, 2020.

\bibitem{asymvalley}
Haowei He, Gao Huang, and Yang Yuan.
\newblock Asymmetric valleys: Beyond sharp and flat local minima.
\newblock In {\em NeurIPS}, 2019.

\bibitem{resnet}
Kaiming He, Xiangyu Zhang, Shaoqing Ren, and Jian Sun.
\newblock Deep residual learning for image recognition.
\newblock In {\em CVPR}, 2016.

\bibitem{ofd}
Byeongho Heo, Jeesoo Kim, Sangdoo Yun, Hyojin Park, Nojun Kwak, and Jin~Young
  Choi.
\newblock A comprehensive overhaul of feature distillation.
\newblock In {\em ICCV}, 2019.

\bibitem{ab}
Byeongho Heo, Minsik Lee, Sangdoo Yun, and Jin~Young Choi.
\newblock Knowledge transfer via distillation of activation boundaries formed
  by hidden neurons.
\newblock In {\em AAAI}, 2019.

\bibitem{kd}
Geoffrey Hinton, Oriol Vinyals, and Jeff Dean.
\newblock Distilling the knowledge in a neural network.
\newblock In {\em arXiv:1503.02531}, 2015.

\bibitem{minima2}
Sepp Hochreiter and J{\"u}rgen Schmidhuber.
\newblock Simplifying neural nets by discovering flat minima.
\newblock In {\em NeurIPS}, 1994.

\bibitem{minima3}
Pavel Izmailov, Dmitrii Podoprikhin, Timur Garipov, Dmitry Vetrov, and
  Andrew~Gordon Wilson.
\newblock Averaging weights leads to wider optima and better generalization.
\newblock {\em UAI}, 2018.

\bibitem{understandingkdji}
Guangda Ji and Zhanxing Zhu.
\newblock Knowledge distillation in wide neural networks: Risk bound, data
  efficiency and imperfect teacher.
\newblock In {\em NeurIPS}, 2020.

\bibitem{NEURIPS2021_892c91e0}
Zijian Kang, Peizhen Zhang, Xiangyu Zhang, Jian Sun, and Nanning Zheng.
\newblock Instance-conditional knowledge distillation for object detection.
\newblock In {\em NeurIPS}, 2021.

\bibitem{minima1}
Nitish~Shirish Keskar, Dheevatsa Mudigere, Jorge Nocedal, Mikhail Smelyanskiy,
  and Ping Tak~Peter Tang.
\newblock On large-batch training for deep learning: Generalization gap and
  sharp minima.
\newblock {\em arXiv:1609.04836}, 2016.

\bibitem{cifar}
Alex Krizhevsky, Geoffrey Hinton, et~al.
\newblock Learning multiple layers of features from tiny images.
\newblock 2009.

\bibitem{NEURIPS2021_018b59ce}
Jogendra~Nath Kundu, Siddharth Seth, Anirudh Jamkhandi, Pradyumna YM, Varun
  Jampani, Anirban Chakraborty, and Venkatesh~Babu R.
\newblock Non-local latent relation distillation for self-adaptive 3d human
  pose estimation.
\newblock In {\em NeurIPS}, 2021.

\bibitem{li2018visualizing}
Hao Li, Zheng Xu, Gavin Taylor, Christoph Studer, and Tom Goldstein.
\newblock Visualizing the loss landscape of neural nets.
\newblock In {\em NeurIPS}, 2018.

\bibitem{dateback3}
Jinyu Li, Rui Zhao, Jui-Ting Huang, and Yifan Gong.
\newblock Learning small-size dnn with output-distribution-based criteria.
\newblock In {\em Interspeech}, 2014.

\bibitem{dateback4}
Percy Liang, Hal Daum{\'e}~III, and Dan Klein.
\newblock Structure compilation: trading structure for features.
\newblock In {\em ICML}, 2008.

\bibitem{understandingkd2}
David Lopez-Paz, Leon Bottou, Bernhard Scholkopf, and Vladimir Vapnik.
\newblock Unifying distillation and privileged information.
\newblock In {\em ICLR}, 2015.

\bibitem{shufflenetv2}
Ningning Ma, Xiangyu Zhang, Hai-Tao Zheng, and Jian Sun.
\newblock Shufflenet {V}2: {P}ractical guidelines for efficient cnn
  architecture design.
\newblock In {\em ECCV}, 2018.

\bibitem{understandingkd3}
Aditya~Krishna Menon, Ankit~Singh Rawat, Sashank~J Reddi, Seungyeon Kim, and
  Sanjiv Kumar.
\newblock A statistical perspective on distillation.
\newblock In {\em ICML}, 2021.

\bibitem{rkd}
Wonpyo Park, Dongju Kim, Yan Lu, and Minsu Cho.
\newblock Relational knowledge distillation.
\newblock In {\em CVPR}, 2019.

\bibitem{visualization2}
Jeffrey Pennington and Yasaman Bahri.
\newblock Geometry of neural network loss surfaces via random matrix theory.
\newblock In {\em ICML}, 2017.

\bibitem{understandingkd1}
Mary Phuong and Christoph Lampert.
\newblock Towards understanding knowledge distillation.
\newblock In {\em ICML}, 2019.

\bibitem{polyak1964some}
Boris~T Polyak.
\newblock Some methods of speeding up the convergence of iteration methods.
\newblock {\em Ussr computational mathematics and mathematical physics}, 1964.

\bibitem{faster_rcnn}
Shaoqing Ren, Kaiming He, Ross Girshick, and Jian Sun.
\newblock Faster r-cnn: Towards real-time object detection with region proposal
  networks.
\newblock In {\em NeurIPS}, 2015.

\bibitem{fitnets}
Adriana Romero, Nicolas Ballas, Samira~Ebrahimi Kahou, Antoine Chassang, Carlo
  Gatta, and Yoshua Bengio.
\newblock Fitnets: {H}ints for thin deep nets.
\newblock In {\em ICLR}, 2015.

\bibitem{imagenet}
Olga Russakovsky, Jia Deng, Hao Su, Jonathan Krause, Sanjeev Satheesh, Sean Ma,
  Zhiheng Huang, Andrej Karpathy, Aditya Khosla, Michael Bernstein, et~al.
\newblock Image{N}et large scale visual recognition challenge.
\newblock {\em IJCV}, 2015.

\bibitem{mobilenetv2}
Mark Sandler, Andrew Howard, Menglong Zhu, Andrey Zhmoginov, and Liang-Chieh
  Chen.
\newblock Mobilenet{V}2: {I}nverted residuals and linear bottlenecks.
\newblock In {\em CVPR}, 2018.

\bibitem{understandingkd5}
Zhiqiang Shen, Zechun Liu, Dejia Xu, Zitian Chen, Kwang-Ting Cheng, and Marios
  Savvides.
\newblock Is label smoothing truly incompatible with knowledge distillation: An
  empirical study.
\newblock In {\em ICLR}, 2021.

\bibitem{vgg}
K. Simonyan and A Zisserman.
\newblock Very deep convolutional networks for large-scale image recognition.
\newblock In {\em ICLR}, 2015.

\bibitem{understanding0}
Samuel Stanton, Pavel Izmailov, Polina Kirichenko, Alexander~A Alemi, and
  Andrew~G Wilson.
\newblock Does knowledge distillation really work?
\newblock In {\em NeurIPS}, 2021.

\bibitem{understandingkd4}
Jiaxi Tang, Rakesh Shivanna, Zhe Zhao, Dong Lin, Anima Singh, Ed~H Chi, and
  Sagar Jain.
\newblock Understanding and improving knowledge distillation.
\newblock {\em arXiv:2002.03532}, 2020.

\bibitem{crd}
Yonglong Tian, Dilip Krishnan, and Phillip Isola.
\newblock Contrastive representation distillation.
\newblock In {\em ICLR}, 2020.

\bibitem{survey}
Lin Wang and Kuk-Jin Yoon.
\newblock Knowledge distillation and student-teacher learning for visual
  intelligence: {A} review and new outlooks.
\newblock {\em T-PAMI}, 2021.

\bibitem{tfkd}
Li Yuan, Francis~EH Tay, Guilin Li, Tao Wang, and Jiashi Feng.
\newblock Revisiting knowledge distillation via label smoothing regularization.
\newblock In {\em CVPR}, 2020.

\bibitem{at}
Sergey Zagoruyko and Nikos Komodakis.
\newblock Paying more attention to attention: {I}mproving the performance of
  convolutional neural networks via attention transfer.
\newblock In {\em ICLR}, 2017.

\bibitem{zhang2021survey}
Yu Zhang and Qiang Yang.
\newblock A survey on multi-task learning.
\newblock {\em TKDE}, 2021.

\bibitem{zhao2022decoupled}
Borui Zhao, Quan Cui, Renjie Song, Yiyu Qiu, and Jiajun Liang.
\newblock Decoupled knowledge distillation.
\newblock In {\em CVPR}, 2022.

\bibitem{NEURIPS2021_29c0c0ee}
Du Zhixing, Rui Zhang, Ming Chang, xishan zhang, Shaoli Liu, Tianshi Chen, and
  Yunji Chen.
\newblock Distilling object detectors with feature richness.
\newblock In {\em NeurIPS}, 2021.

\end{thebibliography}
}

\clearpage
\section*{A.Appendix}

\subsection*{A.1 Algorithm}
\begin{minipage}{0.45\textwidth}
\centering
\begin{algorithm}[H]
\label{algo_sgd}
\caption{Vanilla Trainer}
{\bf Require:} 

\quad  Learning rate $\gamma>0$, 

\quad  momentum coefficient $0<\mu<1$

\quad  loss function $\mathcal{L}$

{\bf Initialize:} 

\quad  $\boldsymbol{\theta}$, $\boldsymbol{v} \leftarrow 0$
\begin{algorithmic}[1]
\While{$\boldsymbol{\theta}$ not converged} 
\State $\boldsymbol{g} \leftarrow \nabla_{\boldsymbol{\theta}}\mathcal{L}(x,y,\boldsymbol{\phi}; \boldsymbol{\theta})$
\State $\boldsymbol{v} \leftarrow \boldsymbol{g} + \mu \boldsymbol{v}$
\State $\boldsymbol{\theta} \leftarrow \boldsymbol{\theta} - \gamma \boldsymbol{v}$
\EndWhile
\end{algorithmic}
\end{algorithm}
\end{minipage}

\begin{minipage}{0.45\textwidth}
\centering
\begin{algorithm}[H]
\label{algo_dot}
\caption{Distillation-Oriented Trainer}
{\bf Require:} 

\quad Learning rate $\gamma>0$, 

\quad momentum coefficient $0<\mu<1$, 

\quad momentum difference $0<\Delta<1-\mu$, 

\quad  loss functions $\mathcal{L}_{\text{CE}}$, $\mathcal{L}_{\text{KD}}$ and 

\quad corresponding weights $\alpha$, $1-\alpha$. 

{\bf Initialize:} $\boldsymbol{\theta}$, $\boldsymbol{v}_{\text{ce}} \leftarrow 0$, $\boldsymbol{v}_{\text{kd}} \leftarrow 0$
\begin{algorithmic}[1]
\While{$\boldsymbol{\theta}$ not converged}
\State $\boldsymbol{g}_{\text{ce}} \leftarrow \alpha\nabla_{\boldsymbol{\theta}}\mathcal{L}_{\text{CE}}(x,y;\boldsymbol{\theta})$
\State $\boldsymbol{g}_{\text{kd}} \leftarrow (1-\alpha)\nabla_{\boldsymbol{\theta}}\mathcal{L}_{\text{KD}}(x,\boldsymbol{\phi};\boldsymbol{\theta})$
\State $\boldsymbol{v}_{\text{ce}} \leftarrow \boldsymbol{g}_{\text{ce}} + (\mu - \Delta) \boldsymbol{v}_{\text{ce}}$
\State $\boldsymbol{v}_{\text{kd}} \leftarrow \boldsymbol{g}_{\text{kd}} + (\mu + \Delta) \boldsymbol{v}_{\text{kd}}$
\State $\boldsymbol{\theta} \leftarrow \boldsymbol{\theta} - \gamma (\boldsymbol{v}_{\text{ce}} + \boldsymbol{v}_{\text{kd}})$
\EndWhile
\end{algorithmic}
\end{algorithm}

\end{minipage}

\subsection*{A.2 A toy experiment for better understanding}
\begin{figure}[h]
\centering
\begin{tiny}
    \includegraphics[width=.45\textwidth]{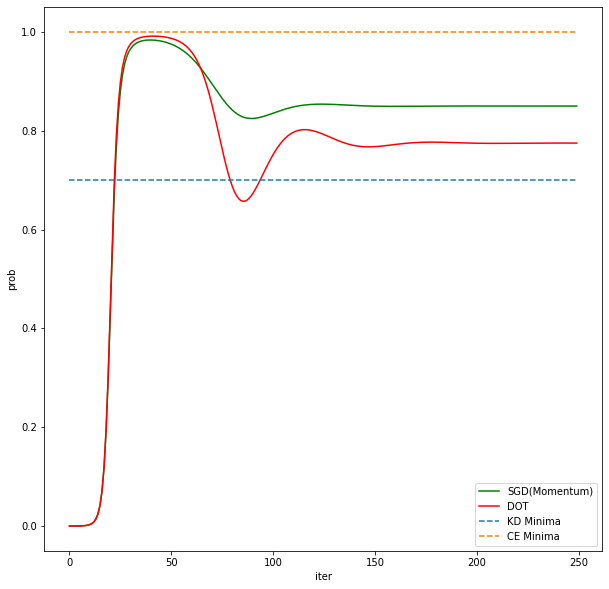}
\end{tiny}
\caption{Toy experiments for analyzing optimization behaviors of SGD and DOT. It conveys that DOT could help the student network converge to minima satisfying the distillation loss well.}
\label{fig:app_toy}
\end{figure}

We conduct a series of \textit{toy} experiments to intuitively illustrate the optimization behaviors of DOT. Concretely, we initialize a 2-d~(trainable) tensor as the logits for a binary classification task. Then, we employ a loss function composed of two parts: (1) a cross-entropy loss~(where the target class is 1), and (2) a distillation loss~(where the teacher's prediction is a constant 0.7). We use a vanilla SGD and our proposed DOT to respectively optimize the loss function, and the prediction of the 2-d tensor is shown in Figure~\ref{fig:app_toy}. It suggests that applying DOT makes the 2-d tensor more similar to the teacher's prediction~(0.7 is the ideal output of a student if distillation loss is well optimized). What's more, DOT could search a wide range of the loss landscape~(great fluctuations in Figure~\ref{fig:app_toy}), which helps the model to get rid of sharp local minima. We hope this toy experiment could provide insights for an intuitive understanding of the working mechanism of DOT.

\subsection*{A.3 More pairs on CIFAR-100}
We conduct more experiments on CIFAR-100 following CRD's protocol, and results are reported in Table~\ref{tab:more_cifar} and~\ref{tab:more_cifar2} which verifies the universality of DOT.

\begin{table}[h]
    \begin{minipage}{.45\textwidth}
    \centering
    \begin{small}
    \begin{tabular}{cc|cc}
        teacher & student & KD & DOT\\
        \hline
        WRN-40-2 & WRN-16-2 & 74.92 & 75.85 \\
        WRN-40-2 & WRN-40-1 & 73.54 & 74.06 \\
        ResNet56 & ResNet20 & 70.66 &  71.07 \\
        ResNet110 & ResNet20 & 70.67 & 71.22 \\
        ResNet110 & ResNet32 & 73.08 & 73.72 \\
        ResNet32$\times$4 & ResNet8$\times$4 & 73.33 & 75.12 \\
        VGG13 & VGG8 & 72.98 & 73.77
    \end{tabular}
    \caption{Pairs of the same architecture.}
    \label{tab:more_cifar}
    \end{small}
    \end{minipage}
    \quad
    \begin{minipage}{.45\textwidth}
    \centering
    \begin{small}
    \begin{tabular}{cc|cc}
        teacher & student & KD & DOT\\
        \hline
        VGG13 & MobileNetV2 & 67.37 & 68.21 \\
        ResNet50 & MobileNetV2 & 67.35 & 68.36 \\
        ResNet50 & VGG8 & 73.81 & 74.38 \\
        ResNet32$\times$4 & ShuffleNetV1 & 74.07 & 74.58 \\
        ResNet32$\times$4 & ShuffleNetV2 & 74.45 & 75.55 \\
        WRN-40-2 & ShuffleNetV1 & 74.83 & 75.92
    \end{tabular}
    \caption{Pairs of the different architectures.}
    \label{tab:more_cifar2}
    \end{small}
    \end{minipage}
\end{table}

\subsection*{A.4 Does longer training time help for better convergence?}
In Section~\ref{sec:sec3} of the manuscript, we visualize and analyze the loss curves and reveal a \textit{trade-off} issue caused by introducing distillation loss. We further conduct the experiment for longer training epochs, \emph{i.e.}, applying a smaller learning rate for extra epochs to study whether the trade-off could be alleviated. Concretely, we further train the network with both task and distillation losses for 60 epochs and decay the learning rate every 30 epochs.
Results in Table~\ref{tab:app_tab1} indicate that longer training still cannot significantly decrease the training task loss. The task loss after longer training is still around 0.38, while the task loss of the vanilla baseline is 0.2379. It indicates that the trade-off issue still remains, further supporting the existence of optimization conflict between task loss and distillation loss.
\begin{table}[h]
    \begin{small}
    \centering
    \begin{tabular}{c|c|ccc}
        epoch & baseline & 240 & 270 & 300  \\ \hline
        validation top-1 & 72.50 & 73.33 & 73.51 & 73.63 \\
        training task loss & \textbf{0.2379} & 0.3844 & 0.3801 & 0.3818
    \end{tabular}
    \caption{Results of training students with both task and distillation losses for longer epochs.}
    \label{tab:app_tab1}
    \end{small}
\end{table}

\subsection*{A.5 Why does DOT perform better on challenging datasets?}
As shown in Table~\ref{tab:main_result}, DOT works better on challenging datasets, \emph{e.g.}, DOT achieves +1$\sim$2\%, 3$\sim$6\% and 1$\sim$2\% performance gain on CIFAR-100, Tiny-ImageNet and ImageNet, respectively. 
We believe the reason is that the teacher could transfer more useful and valuable knowledge on the challenging tasks, and dominating the optimization with distillation loss could better leverage the knowledge, which means the upper bound of the performance gain for DOT is higher.

\subsection*{A.6 About tuning $\Delta$} 
The only hyper-parameter introduced by our DOT is $\Delta$. We notice that the values of $\Delta$ need adjustments on different datasets. However, the improvement is satisfactory without tuning $\Delta$, \emph{i.e.}, $\Delta$ \textit{is not a sensitive hyper-parameter}. Concretely, the value of $\Delta$ for KD+DOT on CIFAR100 is set as 0.075, the for CRD+DOT is set as 0.05, as well as for DKD+DOT. The values of $\Delta$ for all methods on Tiny-ImageNet are set as 0.075. As for ImageNet, knowledge from teachers is more valuable and reliable, so we set $\Delta$ as 0.09 for both KD+DOT and DKD+DOT.

\subsection*{A.7 Implementation of DOT}
It is worth mentioning that the implementation of DOT for feature-based methods is not the same as for logit-based methods~(\emph{e.g.}KD and DKD). The reason is as follows: The extra distillation loss of logit-based methods is a KL-Divergence applied on the student's logits and the teacher's logits, so there are no other extra parameters to optimize and all the parameters of the student network are involved. On the contrary, feature-based methods need extra modules and parameters as connectors between students' features and teachers' features. And the final fully-connected layer~(the classifier) is not involved when computing the gradients of feature-distillation loss. In other words, different losses involve different network parameters in the feature-based distillation methods, so directly applying different momentums for the different losses will lead to a ``gradient inconsistency" problem. To solve this problem, DOT only applies different momentums on the parameters involved by both task and distillation losses. For parameters involved by only one loss~(\emph{e.g.}, the final fully-connected layer), momentums are the same as the baseline.

\end{document}